\documentclass[pmlr]{jmlr}
\usepackage[T1]{fontenc}

\RequirePackage{graphicx}
\usepackage{booktabs}
\usepackage{multirow}
\usepackage{multicol}
\usepackage{longtable}
\usepackage{xcolor}
\usepackage{colortbl}
\usepackage{graphicx}
\usepackage{subcaption}

\definecolor{lightred}{RGB}{245, 204, 204}
\definecolor{lightblue}{RGB}{204, 229, 255}

\makeatletter
\def\set@curr@file#1{\def\@curr@file{#1}} 
\makeatother
\usepackage[load-configurations=version-1]{siunitx} 

\theorembodyfont{\upshape}
\theoremheaderfont{\scshape}
\theorempostheader{:}
\theoremsep{\newline}

\jmlrvolume{298}
\jmlryear{2025}
\jmlrworkshop{Machine Learning for Healthcare}

\newcommand{\datasetname}{\textsc{FactEHR}} 

\title[\datasetname: A Dataset for Evaluating Factuality in Clinical Notes Using LLMs]{\datasetname: A Dataset for Evaluating Factuality in \\ Clinical Notes Using LLMs}

\newcommand{\addrneu}{Khoury College of Computer Sciences, Northeastern University}
\newcommand{\addrcbir}{Center for Biomedical Informatics Research, Stanford University}
\newcommand{\addrdbds}{Department of Biomedical Data Science, Stanford School of Medicine}
\newcommand{\addrshc}{Stanford Health Care}
\newcommand{\addrdm}{Department of Medicine, Stanford School of Medicine}
\newcommand{\addrcerc}{Clinical Excellence Research Center, Stanford School of Medicine}
\newcommand{\addrdappm}{Department of Anesthesiology, Perioperative \& Pain Medicine, Stanford School of Medicine}
\newcommand{\addrdd}{Department of Dermatology, Stanford School of Medicine}
\newcommand{\addrtds}{Technology and Digital Solutions, Stanford Health Care}

\author{\Name{Monica Munnangi*}$^{1,2}$ \Email{munnangi.m@northeastern.edu}
        \AND
        \Name{Akshay Swaminathan*}$^3$ \Email{akshaysw@stanford.edu}
        \AND
        \Name{Jason Alan Fries*\textsuperscript{\dag}}$^2$ \Email{jfries@stanford.edu}
        \AND
        \Name{Jenelle Jindal}$^4$ \Email{jjindal@stanford.edu}
        \AND
        \Name{Sanjana Narayanan}$^2$ \Email{sanjana.books@gmail.com }
        \AND
        \Name{Ivan Lopez}$^3$ \Email{ivlopez@stanford.edu}
        \AND
        \Name{Lucia Tu}$^2$ \Email{ltu@wellesley.edu}
        \AND
        \Name{Philip Chung}$^7$ \Email{chungp@stanford.edu}
        \AND
        \Name{Jesutofunmi A. Omiye}$^{3,8}$ \Email{tomiye@stanford.edu}
        \AND
        \Name{Mehr Kashyap}$^3$ \Email{mkashyap@stanford.edu}
        \AND
        \Name{Nigam Shah}$^{5,6,9}$ \Email{nigam@stanford.edu}
        \AND
        $^1$ \addr \addrneu \\ 
        $^2$ \addr \addrcbir \\
        $^3$ \addr \addrdbds \\
        $^4$ \addr \addrshc \\
        $^5$ \addr \addrdm \\
        $^6$ \addr \addrcerc \\
        $^7$ \addr \addrdappm \\
        $^8$ \addr \addrdd \\
        $^9$ \addr \addrtds \\
}

\begin{document}

\begingroup\def\thefootnote{*}\footnotetext{Equal Contribution}\endgroup
\begingroup\def\thefootnote{\dag}\footnotetext{Corresponding author} \endgroup

\maketitle

\vspace{-2em}
\begin{abstract}

Verifying and attributing factual claims is essential for the safe and effective use of large language models (LLMs) in healthcare. 
A core component of factuality evaluation is \textit{fact decomposition}, the process of breaking down complex clinical statements into fine-grained atomic facts for verification. 
Recent work has proposed fact decomposition, which uses LLMs to rewrite source text into concise sentences conveying a single piece of information, to facilitate fine-grained fact verification.
However, clinical documentation poses unique challenges for fact decomposition due to dense terminology and diverse note types and remains understudied. 
To address this gap and explore these challenges, we present \datasetname, an NLI dataset consisting of document fact decompositions for 2,168 clinical notes spanning four types from three hospital systems, resulting in 987,266 entailment pairs. 
We assess the generated facts on different axes, from entailment evaluation of LLMs to a qualitative analysis. 
Our evaluation, including review by the clinicians, reveals substantial variability in LLM performance for fact decomposition. 
For example, Gemini-1.5-Flash consistently generates relevant and accurate facts, while Llama-3 8B produces fewer and less consistent outputs.
The results underscore the need for better LLM capabilities to support factual verification in clinical text. 
\end{abstract}

\section{Introduction}
\label{sec:introduction}

Verifying and attributing factual claims is essential for the safe and effective use of large language models (LLMs) in healthcare. 
Evaluation strategies have been proposed to assess summarization quality, ensure claims are properly grounded (i.e., attributed to specific text in the clinical note), and reduce hallucinations and related errors \citep{zheng2023judging}. 
A core component of these strategies is \textit{fact decomposition}, which rewrites complex text into concise, atomic statements, each conveying a single piece of information\citep{fabbri-etal-2022-qafacteval, chen-etal-2023-propsegment, min2023factscore}.

Fact decomposition is used to verify facts against source documents using QA systems, textual entailment, or other model-based evaluation methods. 
Textual entailment, in particular, has proven effective for automating factuality assessment \citep{kamoi-etal-2023-wice,ru2024ragcheckerfinegrainedframeworkdiagnosing, xie-etal-2024-doclens}.
Although these verification techniques are well studied in both general and scientific texts \citep{Wadden2020FactOF, Wright2022GeneratingSC}, their application to clinical texts \citep{xie-etal-2024-doclens} remains underexplored. This represents a significant gap, as performance metrics are highly sensitive to the quality of fact decomposition \citep{wanner-etal-2024-closer}.

In healthcare, many documentation tasks require summarizing information from electronic health records (EHRs) \citep{fleming2023medalign}. 
EHRs encompass multiple facets of patient care, including various types of clinical documents, tabular data, and a variety of unstructured data types (e.g., medical imaging, waveforms).
Extracting, rewriting, or verifying evidence from EHRs is necessary for tasks like summarization \citep{Van_Veen_2024, hegselmann2024datacentricapproachgeneratefaithful}, patient phenotyping \citep{yang2024enhancing}, clinical trial recruitment \citep{wornow2024zeroshot}, medical text simplification \citep{devaraj-etal-2021-paragraph}, and knowledge graph construction \citep{arsenyan2023llm}.
Just as facts in radiology reports must be grounded to specific pixel data  \citep{bannur2024maira2groundedradiologyreport}, text generated for documentation tasks must be grounded in data present in the patient's EHR.
Evaluating the ability of LLMs to perform this attribution is necessary for the successful use of LLMs in healthcare.

The decomposition and verification of the facts in clinical notes using LLMs presents significant challenges. Clinical notes employ medical terminology, special acronyms, and non-grammatical shorthand.
Clinical observations are often compositional, as in: ``\textit{lobulation at the apex of the left hemithorax along the mediastinal border is residual of slowly resolving hematoma.}''
For research purposes, clinical notes are frequently stripped of their original markup—like tables, lists, and section headers—thereby increasing ambiguity and parsing difficulties. The de-identification process, which masks patient names and dates, further introduces noise that can confound LLMs. Moreover, medical documentation serves diverse functions: nursing notes offer brief, time-stamped updates on a patient’s physiological status, whereas discharge summaries integrate an entire hospital stay’s details for billing and compliance purposes. Consequently, these documents differ substantially in presentation, length, fact density, and temporal scope.

\begin{figure}[t!]
\centering
  \includegraphics[width=1\columnwidth]{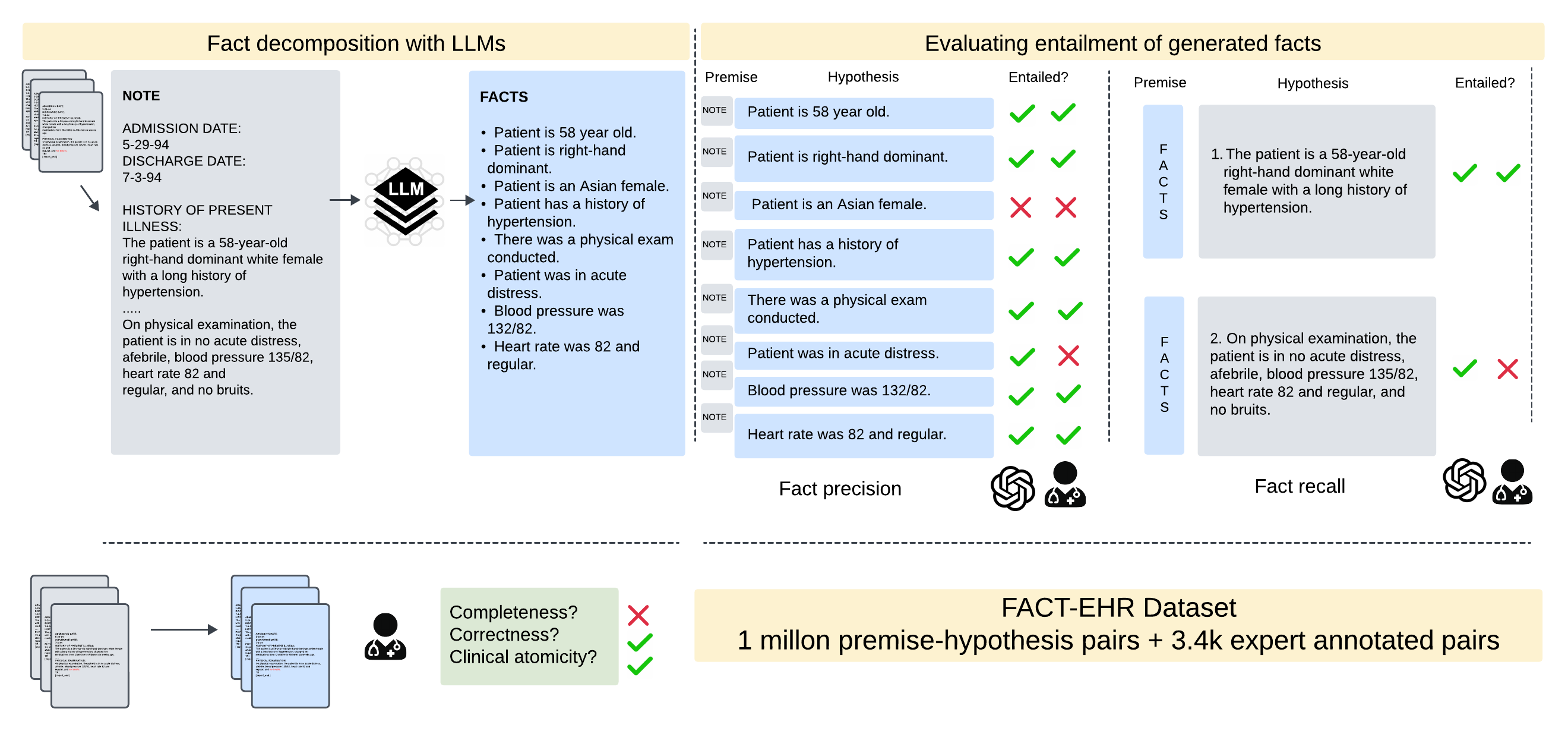}
  \caption{End-to-end pipeline of our work. We generate fact decompositions of clinical notes with LLMs. Evaluate the generated dataset in an entailment setting; For calculating Fact-Precision, we check if the \textbf{generated fact} (the hypothesis) logically follows from original \textbf{clinical note} (the premise) and vice versa for Fact-Recall. We validate the LLM-as-a-Judge with a subset of human evaluation. We also perform a modest amount of qualitative evaluation with experts. }
  \label{fig:pipeline}

\end{figure}

Existing entailment datasets -- which denote whether a piece of text (the \textit{hypothesis}) logically follows from another piece of text (the \textit{premise}) -- do not capture the diversity of clinical documentation, either because they focus on a single note type \citep{miura-etal-2021-improving}, use artificial data \citep{romanov-shivade-2018-lessons}, or have a limited representation of hospital systems. 
Fact decomposition \textit{across} different types of clinical documents has not been evaluated. In addition, it is frequently unclear how to evaluate the quality of a decomposition because determining entailment requires medical expertise. 
Finally, frontier model APIs are often not HIPAA-compliant, hindering evaluations using popular datasets such as MIMIC \citep{Johnson2016MIMICIIIAF}. 

To address this gap and to explore these challenges, we present \datasetname, a fact decomposition and textual entailment dataset derived from 2,168 clinical notes across four document types from three hospital systems. In Figure~\ref{fig:pipeline} provides an overview of the pipeline used to generate fact decompositions, evaluate them using LLM-as-a-Judge (a method that uses a large language model to assess whether a claim is entailed, contradicted, or unsupported by a source text), and validate the results against a subset of expert evaluations.
We also include a small set of qualitative human evaluations. Using this dataset, we explore the following questions and contribute toward answering them:

\paragraph{RQ1: Fact Decomposition} How do fact decompositions vary in length, quality, and similarity across LLMs? We compare outputs from four models and analyze key differences.

\paragraph{RQ2: Entailment} How well can LLMs assess textual entailment in clinical documents? We evaluate performance on existing NLI benchmarks and a new set of expert-labeled entailment pairs derived from clinical notes.

\paragraph{RQ3: Fact Verification} How accurate and complete are LLM-generated fact decompositions of clinical notes? We assess performance using formal metrics and detailed clinical expert review. \\

Our findings reveal considerable variation in the number and atomicity of fact decompositions across clinical documents and across LLMs, with some LLMs generating 2.6x more facts per sentence than others. The quality of fact decompositions also vary highly between LLMs, especially closed models generating factually consistent and covering most information as opposed to smaller open source models, where they are incomplete or they get into infinite generation loops. Consequently, higher fact counts do not correspond to better coverage or consistency—smaller open models often omitted key clinical details or hallucinated, while closed models more reliably captured core information. This raises questions about the validity of metrics that rely on fact decompositions for evaluating LLMs in healthcare documentation tasks. To facilitate future research in this direction, we release code\footnote{\url{https://github.com/som-shahlab/factehr}}, and data\footnote{\url{https://som-shahlab.github.io/factehr-website/}}.


\subsection*{Generalizable Insights about Machine Learning in the Context of Healthcare}

Our study highlights that fact decomposition is a crucial yet overlooked factor in evaluating LLMs for healthcare applications, as variations in decomposition quality significantly impact factuality assessments. We demonstrate that textual entailment, a key method for fact verification, faces unique challenges in clinical settings due to domain-specific complexities like shorthand, compositional observations, and structural inconsistencies. Additionally, we find substantial variability in how different LLMs generate fact decompositions, affecting downstream evaluations and raising concerns about the reliability of existing benchmarks. These insights underscore the need for standardized fact decomposition methodologies and more robust, domain-specific evaluation strategies to ensure accurate and trustworthy use of LLMs in clinical NLP tasks. 

We release a natural language inference (NLI) dataset generated from real clinical notes spanning multiple note types and sources. Fact decompositions are produced by LLMs, with a subset evaluated by medical experts and scaled evaluation provided by LLM-as-a-Judge. 
This dataset can be used as (a) a benchmark for evaluating models on the fact decomposition task and (b) training models using both LLM-generated decompositions and their corresponding evaluations.

\section{Related Work}
\label{sec:background}

\paragraph{Large Language Models (LLMs) in Healthcare}

LLMs encode extensive clinical and biomedical knowledge, recent models specifically, GPT-4 \citep{openai2023gpt4}, as well as Gemini \citep{geminiteam2024gemini15unlockingmultimodal}, have emerged as state-of-the-art LLMs, showcasing impressive capabilities in a wide range of domain-specific applications~\citep{singhal2023large, agrawal2022large, munnangi-etal-2024-fly}. Several evaluations have shown that LLMs can achieve performance comparable to fully supervised models on tasks such as entity extraction and relation extraction \citep{wadhwa2023jointly}, under both few-shot and zero-shot settings.
While considerable progress has been made on atomic fact generation in general domains~\citep{min2023factscore, gunjal2024molecularfactsdesideratadecontextualization} and biomedical literature~\citep{Wadden2020FactOF,Wright2022GeneratingSC}, there is little research focusing on clinical notes. 
Existing frameworks such as MAIRA 2~\citep{bannur2024maira2groundedradiologyreport} address fact-checking in radiology reports but not the broader range of clinical documents. 

\paragraph{Clinical NLI Datasets} Several related NLI datasets have been proposed in the clinical domain.
The NLI4CT dataset focuses on clinical trial reports \citep{jullien-etal-2023-nli4ct} and represents one of the earliest efforts to apply NLI to scientific text, though it is limited in size, containing only 2,400 instances. Another dataset targets social determinants of health extracted from clinical notes \citep{lelkes-etal-2023-sdoh}. Our work differs substantially from both in terms of scale and the diversity of EHR note sources.

\paragraph{LLM-as-a-Judge} Evaluating language model outputs remains an open research challenge, particularly for generative tasks where multiple valid responses exist \citep{chang2023surveyevaluationlargelanguage}. 
In closed-world tasks with a small, discrete set of correct answers—such as classification or multiple-choice question answering—models typically use exact match or likelihood-based selection to produce outputs, which are then evaluated using standard metrics like accuracy.
For short-form text generation, token-overlap metrics such as BLEU \citep{papineni-etal-2002-bleu} and ROUGE \citep{lin-2004-rouge} are commonly used. 
However, these metrics often fail to reward correct but paraphrased outputs. Semantic similarity metrics like BERTScore \citep{zhang2020bertscoreevaluatingtextgeneration} address this limitation by comparing contextualized token embeddings to measure some aspects of meaning, rather than relying on exact lexical overlap.

For complex generative tasks, evaluation is challenging due to the large space of plausible outputs and the potential mixing of correct and incorrect information within a single response. 
Recent work~\citep{min2023factscore, xie-etal-2024-doclens, tian2023finetuning} addresses this by decomposing text into atomic facts to assess factuality and completeness; for example, \citep{xie-etal-2024-doclens} compute claim-level precision and recall using entailment models. 
However, current models often struggle to generate atomic facts that are both accurate and comprehensive, especially in the clinical domain. 
In our experiments, entailment-based evaluation using LLMs showed strong correlation with human judgments. 
Given the high cost and limited scalability of expert annotation, we adopt an LLM-as-a-Judge framework—validated against human evaluations—to assess our dataset.

\section{Methods}
\label{sec:methods}
  
\subsection{Experimental Setup}

We use LLMs to perform fact decomposition, the process of breaking down source text into concise, atomic statements that convey individual pieces of information. We then examine the resulting fact decompositions, comparing their characteristics such as number of decomposed facts, length of the generated facts and similarity across different LLMs. Finally, we pair each decomposition with its corresponding source note to form entailment pairs, evaluating how accurately and completely LLMs capture the original information. In this section, we describe the models, prompts, datasets, types of clinical notes and evaluation, in depth, with more information in the appendix as appropriate.

\paragraph{Models}  We consider four LLMs to perform fact decomposition: GPT-4o \citep{openai2024gpt4ocard}, o1-mini
\citep{singhal2023expertlevelmedicalquestionanswering}, Gemini-1.5-Flash-002 \citep{geminiteam2024gemini15unlockingmultimodal}, and Llama3-8b-Instruct \citep{grattafiori2024llama3herdmodels}. For brevity, we will refer to Llama3-8b-Instruct as Llama3 or Llama3-8B and Gemini-1.5-Flash-002 as Gemini-1.5 in the rest of the paper (unless specified otherwise). We also conducted preliminary experiments with domain specific models including Google’s MedLM but found its performance subpar\footnote{More information in Appendix \ref{apx:models}.}, and so did not pursue further.  All models were run using HIPAA-compliant compute environments and APIs. Model generation hyperparameters are included in Appendix~\ref{sec:hyperparameters}.

\paragraph{Prompts}

We use two distinct prompts: one for fact decomposition and another for entailment evaluation.
The fact decomposition prompt, adapted from \citep{min2023factscore}, includes two in-context examples of clinical note fact decompositions and instructs the LLM to output independent facts as a delimited string (Appendix~\ref{apx:prompts}, Figure~\ref{apx:fact_decomp_prompt}).
The entailment evaluation prompt is tuned using 40 premise-hypothesis pairs sampled from our datasets, optimizing for F1 score. 
The final version, adapted from \citep{xie-etal-2024-doclens}, instructs the LLM to produce a binary entailment judgment in JSON format (Appendix~\ref{apx:entailment_prompt}).

\subsection{Data Sources \& Preprocessing}
\label{ssec:data_souces}


We sample clinical notes from three de-identified research datasets: MIMIC (MIMIC-III \citep{Johnson2016MIMICIIIAF} and MIMIC-CXR \citep{johnson2019mimic}), from Beth Israel Deaconess Medical Center in Boston, MA; CORAL, from the University of California, San Francisco (UCSF) \citep{doi:10.1056/AIdbp2300110}; and MedAlign, from Stanford Health Care (SHC), Palo Alto, CA \citep{fleming2023medalign}.
We randomly sample up to 250 notes per type from each dataset, limiting note lengths to 64–3840 whitespace-delimited tokens, for a total of 2,168 notes as summarized in Table \ref{tab:data_statistics} and Table \ref{tab:tokens} for token length. Additional details are in Appendix \ref{apx:dataset_sources}.

\subsection{Note Types}

We consider four clinical note types: (1) \textit{Procedure Note}, which typically includes procedures, clinical indications, findings, and follow-up recommendations. (2) \textit{Nursing Note}, which provides a systematic assessment of a patient's condition across body systems (e.g., cardiovascular, neurologic) at a specific time, with less emphasis on future care planning. (3) \textit{Progress Note}, which summarizes a patient's medical status from the previous day and outline the care plan for the next. (4) \textit{Discharge Summary}, which provides a concise overview of the patient’s presentation, past medical history, key findings, future medical plans, and discharge medications for subsequent care providers. We provide more details about note types in Appendix~\ref{apx:note_types}.


\subsection{Entailment evaluation (Validating LLM-as-a-Judge)}
\label{ssec:entailment}

This study benchmarks the LLM-as-a-judge approach for fact verification in clinical texts, a previously under-evaluated component. To ensure statistical validity of performance metrics like sensitivity and specificity, power calculations \citep{https://doi.org/10.1111/j.1553-2712.1996.tb03538.x} guided the scale of human annotations based on NLI results from the FactEHR dataset. We annotated 1000 unique entailment pairs (250 per note type) and included 200 duplicates (50 per note type) to assess inter-annotator agreement, totaling 1200 annotations in this tranche—sufficient for high-confidence estimates (99\% overall, ~80\% per note type). Combined with the 2468 previously annotated pairs, our dataset now comprises ~3500 human-annotated clinical entailment pairs, making it one of the largest of its kind in the domain. 

To evaluate the accuracy and completeness of fact decompositions, we pair each decomposition with its corresponding source note to form premise–hypothesis pairs for textual entailment. We assess whether LLMs can reliably evaluate entailment in this setting (see Appendix~\ref{apx:supp_res}, Table~\ref{tab:validating_llm_as_a_judge} for details). As part of early model development, we also benchmarked prior models such as RoBERTa, finding that their performance was consistently lower than that of frontier LLMs like GPT-4o.
Additionally, we explored entity-level metrics such as RadGraph F1 \citep{jain2021radgraphextractingclinicalentities}, but found they did not correlate well with expert judgments of factuality in our domain, highlighting the need for more nuanced and holistic evaluation approaches.

We first benchmark five LLMs (GPT-4o, GPT-4o-mini, Gemini-1.5, Llama3-8B, and Llama3-70B) using four publicly available NLI benchmarks: SciTail \citep{DBLP:conf/aaai/KhotSC18}, MedNLI \citep{romanov-shivade-2018-lessons}, MultiNLI \citep{williams-etal-2018-broad}, and SNLI \citep{bowman-etal-2015-large}. Additional information about these datasets is provided in Appendix~\ref{apx:NLI_datasets}. We also evaluate performance on the FactEHR development set, which is described below.
All models use the entailment prompt shown in Figure~\ref{apx:entailment_prompt}, applied to all 987,266 entailment pairs in the FactEHR dataset. Of these, 1,036 pairs were manually annotated by clinical experts, enabling direct assessment of model accuracy.

\subsection{Fact Verification}
\label{ssec:metrics}

To assess the accuracy and completeness of model-generated fact decompositions, we use \textit{fact-precision} and \textit{fact-recall}, entailment-based metrics proposed by DocLens \citep{xie-etal-2024-doclens}. \\

Let \( S \) denote the set of sentences obtained by tokenizing the clinical note \( d \), and let \( C \) represent the set of factual claims produced by decomposing \( d \). We define \([d \models c] = 1\) if a hypothesis \( c \) is completely entailed by a premise \( d \), and \( 0 \) otherwise. For example, the hypothesis, \textit{``There is a pleural effusion on the left side,''} is completely entailed by the premise \textit{``Left pleural effusion increased from prior scan.''} 

\paragraph{Fact Precision} 
Fact precision measures the accuracy of decomposed claims using entailment, treating each fact as a hypothesis and the entire clinical note as the premise. It is defined as the proportion of facts in \( C \) that are entailed by clinical note \( d \):
\(P^{*} = \frac{1}{|C|} \sum_{c} [d \models c]\)

\paragraph{Fact Recall} 
Fact recall measures the completeness of a fact decomposition using entailment, treating each sentence of the clinical note as a hypothesis and the entire set of facts as the premise. It is defined as the proportion of sentences in \( S \) that are entailed by the fact decomposition \( C \):
\[
R{\text{*}} = \frac{1}{|S|} \sum_{s \in S} [C \models s],
\]
where \( S \) is the set of sentences in the clinical note, and \( [C \models s] \) is an indicator function that evaluates to 1 if \( s \) is entailed by \( C \), and 0 otherwise.

We use a weighted fact recall metric to account for multiple facts within a single sentence and parsing issues. Instead of counting each sentence as a single fact, we estimate the number of facts per sentence. Sentences that contain incomplete or invalid information (e.g., just units, numbers, or parsing errors) receive a weight of zero. All other sentences are weighted by their estimated fact count, determined by the sentence atomicity measure described earlier. The weighted fact recall is then defined as:
\[
R{\text{*}}_w = \frac{\sum_{s \in S} w_s [C \models s]}{\sum_{s \in S} w_s},
\]
where \( w_s \) is the weight assigned to sentence \( s \), representing the estimated number of facts it contains.

\section{\datasetname}
\label{sec:dataset}

%

\begin{figure*}[th!]
    \centering
    {\includegraphics[width=\textwidth]{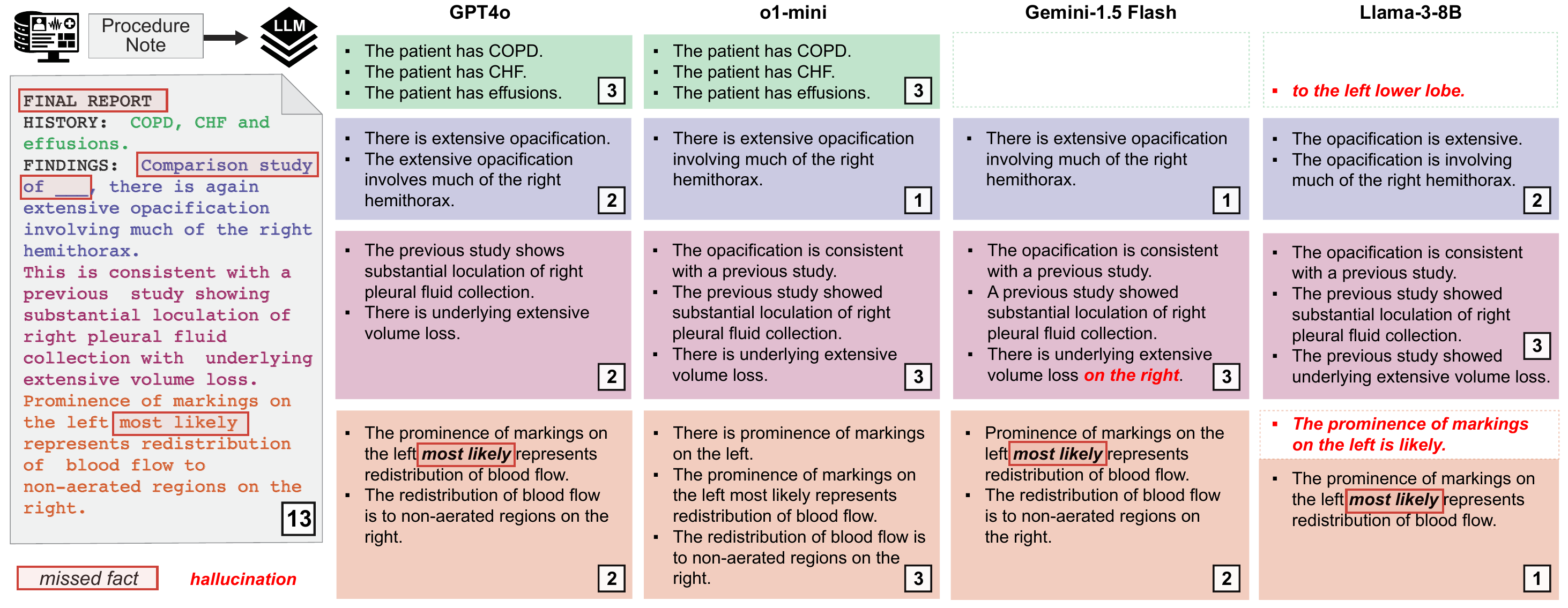}}
    \caption{Example fact decompositions across LLMs. Colors indicate mappings from note (left) to corresponding facts, numbered boxes indicate fact counts, and red text indicates hallucination, red boxes indicated missed facts. All LLMs missed the the comparison study (a prior radiograph) and the note's final report status.} 
    \label{fig:fact_decomp_llms}
\end{figure*}

Complete details on data sources and preprocessing are provided in Subsection~\ref{ssec:data_souces}.
Figure~\ref{fig:fact_decomp_llms} shows example fact decompositions from each LLM evaluated in this study.
Table~\ref{tab:named_subsets} enumerates key dataset summary statistics.

\subsection{Data Generation} 

We introduce \datasetname~(\textit{pronounced ``factor''}), a dataset for fact decomposition and textual entailment created from de-identified clinical notes originating from three hospital systems. \datasetname~includes:

\begin{itemize} \item \textbf{Fact Decompositions}: 8,665 fact decompositions produced by four LLMs from 2,168 clinical notes spanning four document types. \item \textbf{Entailment Pairs}: 987,266 entailment pairs generated by pairing fact decompositions with their source text. Each pair has a binary entailment label from GPT-4o, and a subset of 1,036 pairs also includes labels from clinical experts. \item \textbf{Development Set}: A separate set of 2,468 entailment pairs annotated by clinical experts, used for entailment model development and tuning. \end{itemize}


\begin{table}[th!]
\centering
\small
\begin{tabular}{l@{\hskip 8pt}l@{\hskip 8pt}r@{\hskip 8pt}p{5.2cm}}
\toprule
\textbf{Name} & \textbf{Creation Rule} & \textbf{Size} & \textbf{Description} \\
\midrule
\texttt{FactNotes}      
& \texttt{note $\in$ D} 
& 2,168     
& A collection of clinical notes from the dataset D. \\

\texttt{FactDecomp}     
& \texttt{note $\rightarrow$ fact-list} 
& 8,665 
& Decomposition of notes into a list of structured, atomic facts. \\

\texttt{FactEntail\_p}  
& \texttt{I[note $\Rightarrow$ fact]} 
& 491,663   
& Entailment pairs for evaluating whether a note implies a given fact. \\

\texttt{FactEntail\_r}  
& \texttt{I[fact-list $\Rightarrow$ sentence]} 
& 495,603 
& Entailment pairs for evaluating whether a fact list implies a full sentence. \\

\texttt{FactEntail}     
& \texttt{FactEntail\_p $\cup$ FactEntail\_r} 
& 987,266 
& Union of both entailment settings to support broad NLI evaluation. \\

\texttt{FactEntail\_ann} 
& \texttt{x $\sim$ U(FactEntail)} 
& 1,036 
& Human-annotated entailment data sampled from \texttt{FactEntail\_p} and \texttt{FactEntail\_r}. \\
\bottomrule
\end{tabular}
\caption{\datasetname~dataset overview. We generate fact decompositions from 2,618 clinical notes and derive multiple NLI subsets.}
\label{tab:named_subsets}
\end{table}

\paragraph{Sampling} FactEHR includes 8,665 fact decompositions generated by four LLMs from 2,168 clinical documents (4 x 2,168 = 8,172; five decompositions were not produced due to content moderation policies). To evaluate fact precision and recall, we formed 987,266 pairs from these decompositions. Each pair is classified as either a fact-precision pair, where the premise is the clinical note and the hypothesis is a fact from the decomposition, or a fact-recall pair, where the premise is the fact decomposition and the hypothesis is a sentence from the corresponding clinical note. 

\paragraph{Entailment annotation} From these 987,266, we randomly selected 1,036 pairs for manual entailment annotation by clinical experts. Clinical experts included two board certified physicians, two residents, two medical students, and a clinical researcher. Annotators were provided instructions for labeling entailment (Appendix \ref{apx:annotation_guidelines}), and a subset of 100 entailment pairs were labeled in duplicate to calculate inter-rater agreement.

\section{Results}
\label{sec:results}

\subsection{RQ1 — Fact Decomposition}

We examine the length and similarity of fact decompositions across LLMs, as well as the atomicity of source sentences.

\paragraph{Fact Counts}
\label{par:fact_dispersion}
We report the average number of facts generated per source note sentence across models and note types. To assess the variation in the number of facts generated across LLMs, we compute the coefficient of variation (standard deviation divided by the mean) of the number of facts generated by each LLM for each document \citep{everitt2006cambridge} of these counts. 
This metric, calculated across all documents within each note type, quantifies the degree of disagreement among LLMs on the total number of facts per document.

For all note types, GPT-4o and o1-mini produce more facts per sentence than Gemini-1.5 and Llama3-8B (Table \ref{tab:fact_count}). For discharge summaries, o1-mini generates 1.55 facts per sentence compared to 0.98 for Gemini-1.5 and 0.60 for Llama3-8B.

\begin{figure}
\centering
\begin{minipage}{0.48\textwidth}
  \centering
  \includegraphics[width=0.8\linewidth]{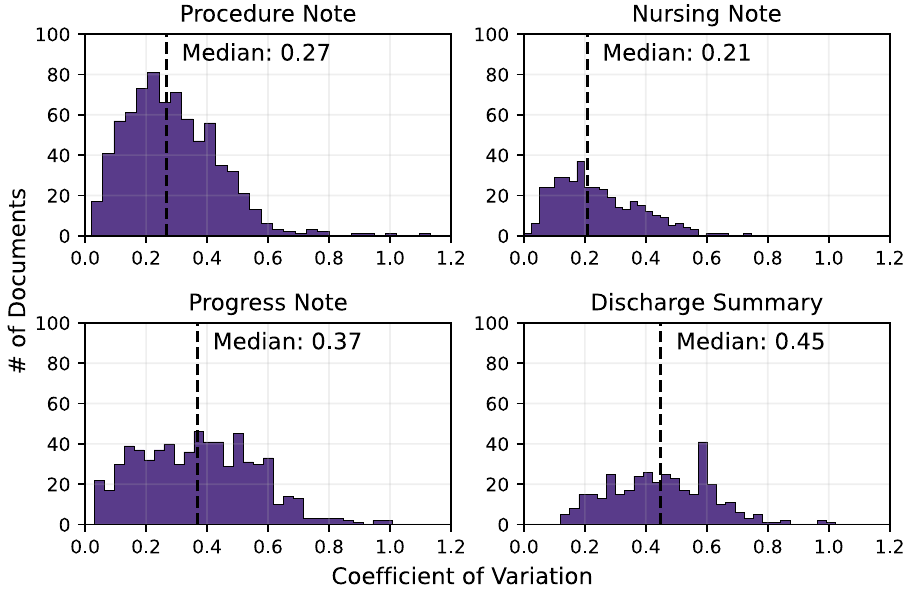}
  \caption{\small Distribution of coefficient of variation across documents. Higher values indicate higher variance in number of facts across LLMs.}
  \label{fig:coeff_plot}
\end{minipage}%
\hspace{0.2em}
\begin{minipage}{0.48\textwidth}
  \centering
  \includegraphics[width=0.8\linewidth]{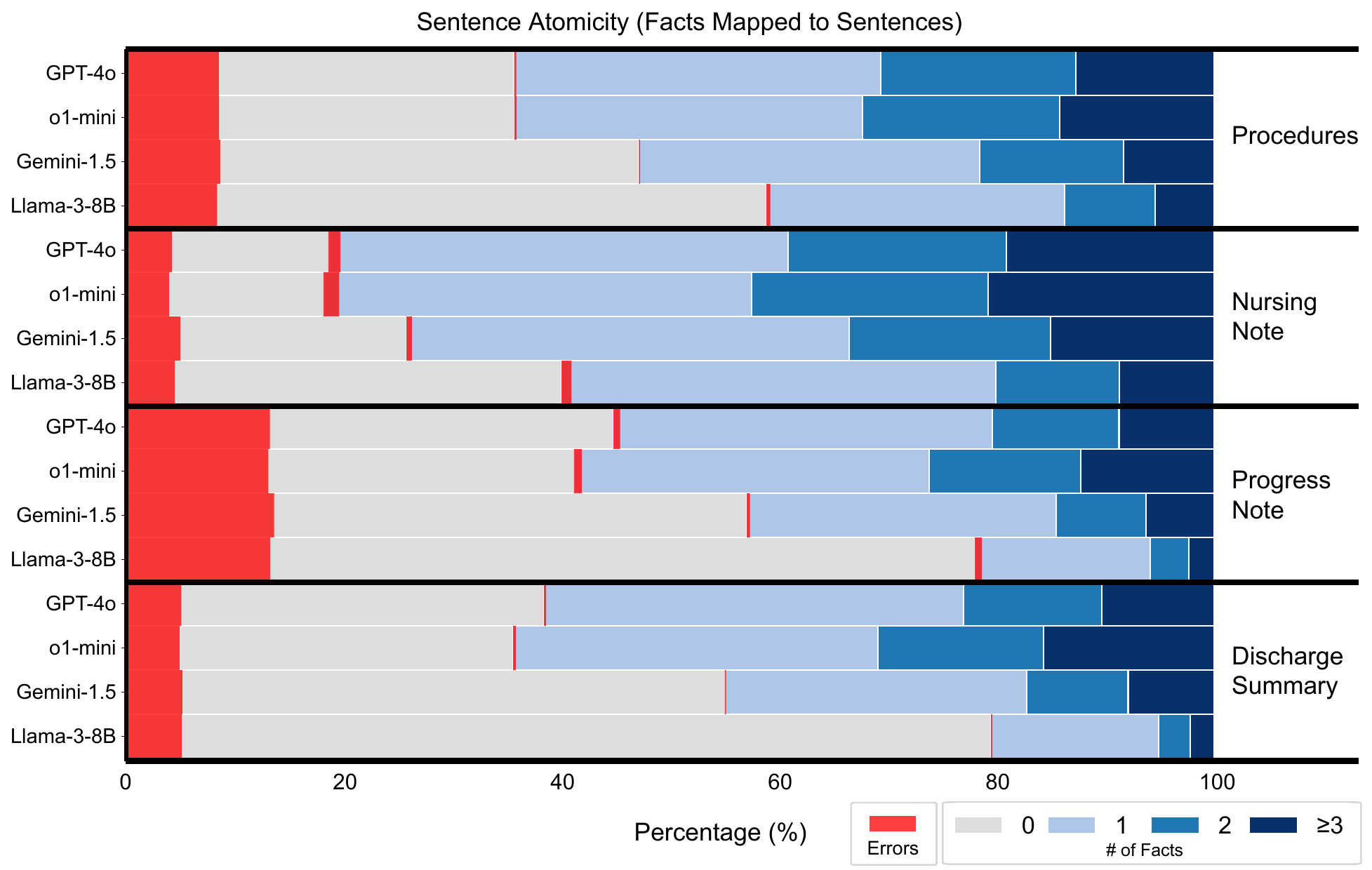}
  \caption{\small Sentence atomicity by note type and LLM, shown as the normalized distribution of the estimated number of facts-per-sentence}
  \label{fig:sent_atomicity}
\end{minipage}
\end{figure}

Figure~\ref{fig:coeff_plot} illustrates the distribution of the coefficient of variation (CV) in generated facts across clinical note types, showing the variability in the number of facts generated per document by LLMs. Discharge summaries exhibited the highest median CV (0.45), followed by progress notes (0.37), procedure notes (0.27), and nursing notes (0.21). The lower variability observed in procedure and nursing notes is expected given shorter token lengths.

\paragraph{Fact Similarities} To quantify the similarity of fact decompositions across LLMs, we use Earth Mover's Distance (EMD) \citep{rubner1998metric, kusner2015word}, which calculates the minimum cost to transform one probability distribution into another. Specifically, we embed each fact using ClinicalBERT \citep{alsentzer-etal-2019-publicly} and represent a single decomposition as a matrix of these fact embeddings. When comparing two decompositions, one serves as the source distribution of fact embeddings and the other as the target distribution. We then compute EMD using a cosine cost function and uniform weighting to determine how closely the source’s set of facts aligns with the target’s set, yielding a measure of similarity between the two decompositions.

\begin{figure*}[!t]
    \centering
    {\includegraphics[trim={0.2cm 0 0.8cm 0},clip,width=\textwidth]{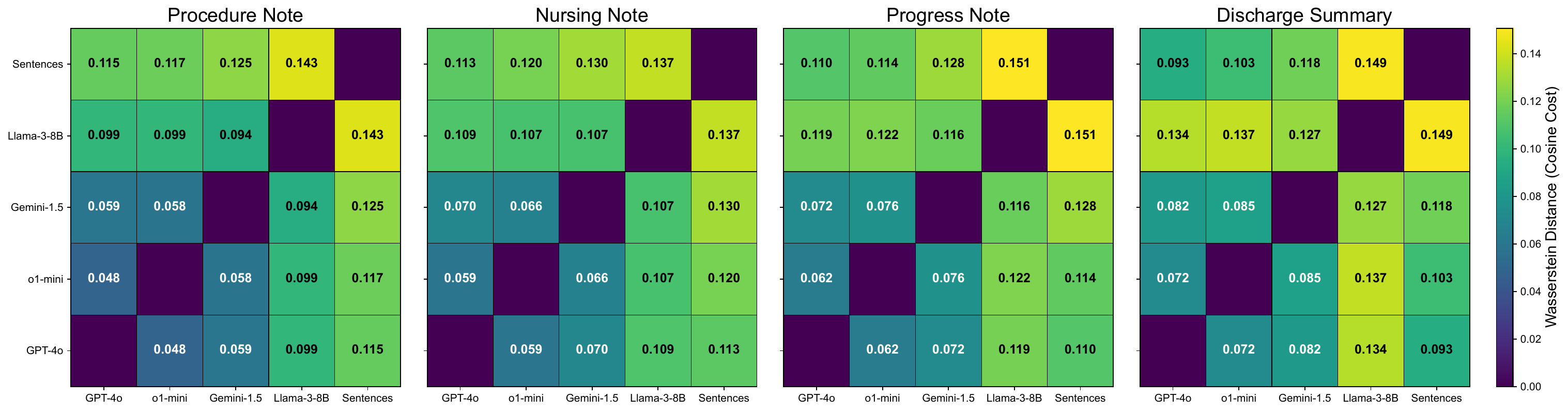}}
    \caption{Mean EMD (cosine cost) across LLM fact decompositions and source sentences. This measures how embedded fact decompositions vary across LLMs and in comparison to the original document.} 
    \label{fig:emd}
\end{figure*}

Using (EMD) with a cosine cost function, we found that GPT-4o and o1-mini produced the most similar fact decompositions, closely followed by Gemini-1.5, whose pairwise EMD values against these models were consistently within 0.02 units of each other (Figure \ref{fig:emd}). In contrast, Llama3’s decompositions were notably more divergent, with pairwise EMD scores roughly 0.03–0.05 units higher. Compared to the original source sentences, GPT-4o decompositions had the lowest mean EMD, while Llama3’s were highest. Differences were also influenced by note type: procedure notes showed minimal variation due to their structured format, nursing notes had slightly higher EMDs reflecting more dynamic content, and discharge summaries—being lengthy and complex—exhibited the highest EMD values.

\paragraph{Sentence Atomicity} To quantify the amount of information in each sentence of a source document, we estimate the number of  facts contained within a single sentence, which we refer to as sentence atomicity. 
We assign each fact to its most similar sentence using a greedy matching approach based on cosine similarity of ClinicalBERT embeddings.

\label{ssec:result_rq1}

Overall, o1-mini produced the most facts per sentence, followed closely by GPT-4o, both maintaining similar rates of zero-fact sentences (Figure~\ref{fig:sent_atomicity}). Gemini-1.5 had more zero-fact sentences than GPT-4o and o1-mini, but fewer than Llama3.

\begin{table}[h!]
\centering
\small
\renewcommand{\arraystretch}{1.2}
\setlength{\tabcolsep}{6pt}
\begin{tabular}{l c c c c}
\toprule
\textbf{Model Name} & \textbf{Procedure Note} & \textbf{Nursing Note} & \textbf{Progress Note} & \textbf{Discharge Summary} \\
\midrule
GPT-4o      & 1.45 $\pm$ 0.03 & 2.37 $\pm$ 0.14 & 1.28 $\pm$ 0.03 & 1.30 $\pm$ 0.02 \\
o1-mini     & 1.45 $\pm$ 0.03 & 2.42 $\pm$ 0.13 & 1.46 $\pm$ 0.03 & 1.55 $\pm$ 0.03 \\
Gemini-1.5  & 1.12 $\pm$ 0.03 & 2.05 $\pm$ 0.12 & 1.02 $\pm$ 0.02 & \cellcolor{lightred}0.98 $\pm$ 0.02 \\
Llama3 8B   & \cellcolor{lightred}0.89 $\pm$ 0.03 & 1.58 $\pm$ 0.12 & \cellcolor{lightred}0.73 $\pm$ 0.04 & \cellcolor{lightred}0.60 $\pm$ 0.04 \\
\bottomrule
\end{tabular}
\caption{Average facts per sentence by model and note type. Red indicates fewer facts than sentences, suggesting the LLM may fail to capture all facts.}
\label{tab:fact_count}
\end{table}

Llama3 produced facts for only about 20\% of sentences in progress notes and discharge summaries, and, except for nursing notes, generated fewer facts than sentences (Table~\ref{tab:fact_count}).

Nursing notes have the highest proportion of sentences with three or more facts, reflecting their high information density. Progress notes, which tend to be longer and more narrative, have the lowest proportion of such sentences.

\begin{table*}[t]
\centering
\small
\begin{tabular}{l|rrrrr|rrrrr}
\toprule
 & \multicolumn{5}{c}{Procedure Note} & \multicolumn{5}{c}{Nursing Note} \\
 &$ P* $ & $ R* $ & $ R*_w $ & $ F1* $ & $ F1*_w $ & $ P* $ & $ R* $ & $ R*_w $ & $ F1* $ & $ F1*_w $ \\
\midrule
GPT-4o & \textbf{98.5} & \textbf{78.7} & 92.4 & \textbf{86.5} & 95.0 & \textbf{97.5} & \textbf{88.9} & 92.2 & \textbf{92.3} & 94.2 \\
o1-mini & 97.8 & 78.4 & \textbf{93.0} & 86.2 & \textbf{95.4} & 96.6 & 88.5 & \textbf{93.3} & 91.6 & \textbf{94.4} \\
Gemini-1.5 & 95.9 & 64.2 & 85.7 & 77.0 & 91.7 & 93.2 & 77.3 & 83.7 & 84.4 & 88.3 \\
Llama3-8B & 84.2 & 49.4 & 73.7 & 62.0 & 82.3 & 84.1 & 56.6 & 69.9 & 65.1 & 75.6 \\
\bottomrule
\end{tabular}

\begin{tabular}{l|rrrrr|rrrrr}
\toprule
 & \multicolumn{5}{c}{Progress Note} & \multicolumn{5}{c}{Discharge Summary} \\
 & $ P* $ & $ R* $ & $ R*_w $ & $ F1* $ & $ F1*_w $ & $ P* $ & $ R* $ & $ R*_w $ & $ F1* $ & $ F1*_w $ \\
\midrule
GPT-4o & \textbf{97.4} & 78.1 & 85.2 & 85.8 & \textbf{89.9} & \textbf{97.0} & 79.0 & 86.9 & \textbf{86.8} & \textbf{91.1} \\
o1-mini & 95.2 & \textbf{79.7} & \textbf{85.7} & \textbf{86.6} & 89.7 & 94.8 & \textbf{81.1} & \textbf{88.2} & 86.7 & 90.0 \\
Gemini-1.5 & 96.0 & 66.2 & 75.2 & 77.9 & 83.9 & 96.5 & 65.4 & 76.3 & 76.9 & 83.8 \\
Llama3-8B & 83.4 & 41.9 & 57.4 & 53.8 & 67.0 & 84.6 & 38.9 & 57.3 & 50.4 & 67.2 \\
\bottomrule
\end{tabular}
\caption{Comparison of average \textit{fact precision} (P$^{*}$), unweighted (R$^{*}$) and weighted (R$^{*}_{w}$) \textit{fact recall} and unweighted (F1$^{*}$) and weighted (F1$^{*}_{w}$) \textit{fact F1} across documents for each note type and fact decomposition model. \textbf{Bolded} are the highest numbers (F1 and accuracy) of a model for each dataset. GPT-4o and o1-mini emerge as top performers across all note types.}
\label{tab:main_results}
\vspace{-1em}
\end{table*}

In summary, quantifying this variability is relevant for two reasons: (1) Fact Verification Approaches: Current workflows use LLM-generated decompositions to break claims into sub-claims for verification. However, our analysis reveals inconsistencies across LLMs. Ideally, facts should be atomic, addressing one property or action at a time. When decompositions fail to achieve this, they introduce uncertainty and lead to varying conclusions, and (2) Clinical Document Types: Clinical documents vary in length, structure, and complexity, challenging LLM-based decomposition. Radiology reports are concise, while discharge summaries are lengthy and complex. Procedure notes require detailed step-by-step breakdowns. LLMs struggle with longer, intricate documents, often producing incomplete decompositions. Addressing these challenges is key to improving LLM performance in clinical documentation.

\subsection{RQ2 — Entailment}
GPT-4o achieve the best F1 scores on four of the five benchmark datasets, including the FactEHR development set in table~\ref{tab:NLI_benchmarks} (Appendix~\ref{apx:supp_res}), and was therefore selected as the entailment judge for all 987,266 entailment pairs in the FactEHR dataset. From these, we randomly sampled 1,036 pairs for manual annotation by seven clinical experts. On doubly annotated entailment pairs, Fleiss' Kappa was 0.67, indicating substantial inter-rater agreement. Under this evaluation, GPT-4o achieves a recall of 0.96 and precision of 0.86 (Appendix~\ref{apx:supp_res}, Table \ref{apx:entail_valid}), closely mirroring its performance on the FactEHR development set. Overall, these findings suggest that GPT-4o serves as a reasonable proxy for human judgment in entailment assessments.

\subsection{RQ3 — Fact Verification} 

Using GPT-4o as the LLM entailment judge, we compute fact precision, fact recall (weighted and unweighted), and fact F1 (weighted and unweighted) for each fact decomposition across all note types (Table~\ref{tab:main_results}). GPT-4o and o1-mini emerge as top performers across all note types. For example, GPT-4o achieves a slightly higher fact F1 than o1-mini on procedure notes (86.5 vs. 86.2) and nursing notes (92.3 vs. 91.6), while o1-mini outperforms GPT-4o on progress notes (86.6 vs. 85.8). On discharge summaries, the two models are nearly tied (86.8 for GPT-4o vs. 86.7 for o1-mini). When considering the weighted fact F1 metric, o1-mini surpasses GPT-4o on procedure notes and nearly closes the gap on nursing notes. This suggests that the weighting scheme, which accounts for differences in fact counts per note, can influence which model is deemed superior.

Across all note types, fact precision remained consistently higher and less variable than fact recall—indicating that while the generated facts were generally accurate, many relevant facts were omitted, resulting in greater variability in recall and aligning with our qualitative findings (Appendix~\ref{apx:supp_res}, Figure~\ref{fig:prec_rec_variance}).


\paragraph{Qualitative Expert Evaluation} Expert reviewers assessed overall completeness of fact decompositions and the correctness, independence, and atomicity of individual facts. Completeness, defined as capturing all information in the source, is low: only 5\% (1/20) of decompositions meet this criterion. For individual facts, reviewers judge whether each fact accurately reflected the source (correctness), stood on its own (independence), and represented a minimal unit of information (atomicity), we present detailed annotation guidelines in Appendix \ref{apx:qual_anno_guidelines}.

\begin{figure*}[t]
    \centering
    {\includegraphics[trim={0 3.5cm 0 4cm},clip,width=\textwidth]{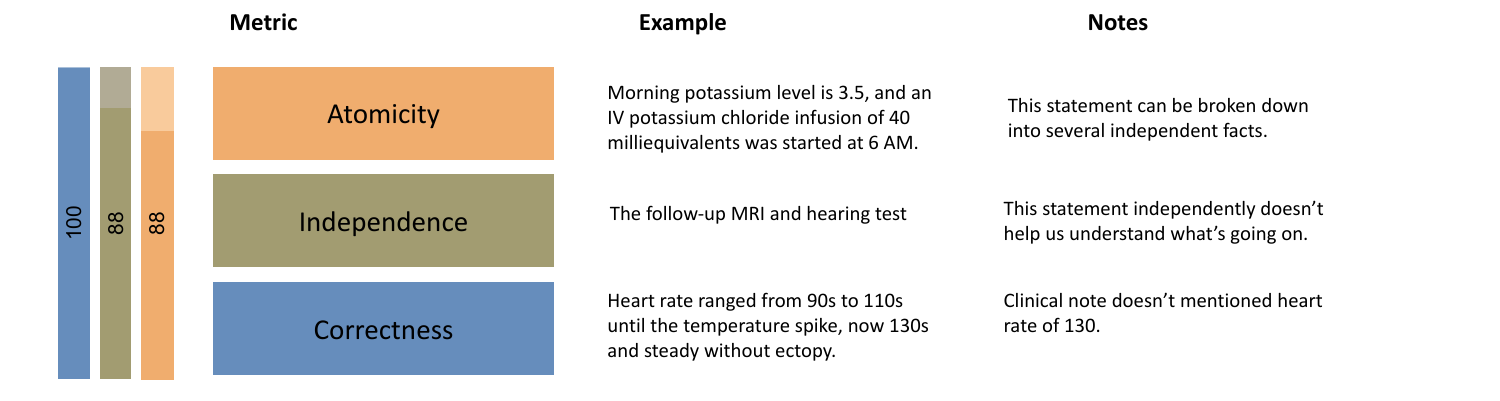}}
    \caption{Overview of in-depth qualitative review with \textbf{GPT-4o} on the fact decomposition on twenty randomly-selected examples from FactEHR. We report \textbf{percentage} of correct, independent and atomic facts as annotated by medical expert.} 
    \label{fig:qual_eval_gpt4}

\end{figure*}

Across models, Llama3 had the highest rate of incorrect facts. Discharge summaries yield the least number of atomic facts, reflecting their complexity, while progress and procedure notes contain more granular and independent facts. Detailed results for GPT-4o are shown in Figure~\ref{fig:qual_eval_gpt4}; additional results for other models and a summary of key metrics are reported in Appendix~\ref{apx:other_qual_results} and Table~\ref{tab:qual_results_summary}, respectively.

\begin{table}[h!]
\centering
\begin{small}
\begin{tabular}{lccc}
\toprule & 
 \textbf{Correct} & \textbf{Independent} & \textbf{Atomic} \\
\midrule
GPT-4o & 100 & 88 & 88   \\
Gemini-1.5 & 96 & 94 & 81  \\
Llama3-8B & 45 & 95 & 74  \\
\bottomrule
\end{tabular}
\end{small}
\caption{Overview of qualitative (fact-level) review on the fact decomposition for 20 notes. We report \textbf{percentage} of correct, independent and atomic facts as annotated by medical expert.}
\label{tab:qual_results_summary}
\end{table}

Overall, fact decompositions by Llama3 on long notes are subpar, specifically they are either incomplete or they get into infinite generation loops. Fact decompositions by GPT-4o are better, but also not complete when there are long notes. Fact decomposition appears to be easier for shorter notes than super long/dense notes.  Even though some long notes get decomposed to 200+ facts, they are still missing content that was not decomposed.
Llama3 and GPT-4o seem to have a greater tendency to produce incomplete statements which is measured as a non-independent statement as compared to Gemini and o1-mini.
\section{Discussion} 
\label{sec:discussion}

In this study, we evaluate four LLMs’ ability to decompose over 2,000 real-world, EHR-derived clinical notes of four types into independent facts. We evaluate the fact decompositions with an LLM-as-a-Judge (validated by expert evaluation), as an entailment task. For calculating Fact-Precision, we check if the generated fact (the hypothesis) logically follows from original
clinical note (the premise) and vice versa for Fact-Recall\footnote{Detailed explanation in Section~\ref{ssec:metrics}}. To enable reproducibility and further research in this direction, we will release \datasetname, a dataset encompassing 8,665 fact decompositions and 987,266 entailment pairs from four note types and three institutions, including 1,036 pairs annotated by seven clinical experts. The dataset was specifically designed to address a gap in the literature regarding the accuracy of fact decomposition in clinical text. While fact decomposition is a critical first step for many fact-checking and verification methods, such as FactScore \citep{min2023factscore}, there is limited research on its effectiveness, particularly when applied to both human- and LLM-generated clinical text. Our work aims to fill this gap by evaluating LLM performance on this foundational task, which is often overlooked in existing studies. Additionally, we examine the length and similarity of the decompositions and find that there was up to a 2.6-fold difference in the number of facts generated by different LLMs. Clinician review confirmed that while the generated facts were generally accurate, although, important details were often omitted.

While choosing the LLMs for fact decomposition task,  we focused on evaluating a representative set of high-performing general-domain models, as these have demonstrated broad utility and scalability for real-world applications. That said, we acknowledge the importance of benchmarking against additional domain-specific models and discuss the need to evaluate them in future work. We also emphasize that the methodology proposed in this study is model-agnostic and can be applied to evaluate the performance of other models, including ones that are medically pretrained. The release of the FactEHR dataset directly supports future evaluation of LLMs by the community.

In the future, we hypothesize that the dataset could be used to fine-tuning a smaller language model for fact-decomposition tasks, which could leverage the training and human-evaluated data splits from \datasetname~to achieve higher quality decompositions tailored to specific needs. 
The results highlight the need to improve LLM-based fact decomposition methods for clinical documents to support the use of LLMs for healthcare use.

\paragraph{Limitations}
\label{sec:limitations}
  
There are several limitations to this work. First, our study focuses on four common note types, but many other specialized note types (e.g., ophthalmology) may yield even greater variability in fact decomposition, suggesting our error estimates are conservative. Second, we rely on a combination of open-source and proprietary models within a HIPAA-compliant environment, ensuring privacy but limiting reproducibility; we mitigate this by releasing our dataset, prompts and code. Third, lack of access to ground truth fact decompositions for clinical notes. This constraint forces us to rely on estimates and proxy measures, such as Fact-Recall, to assess the completeness of model-generated fact decompositions. These metrics are inherently limited in their ability to capture the true completeness and accuracy of the decompositions. Lastly, since we examine only English documents, the generalizability of these findings to other languages remains uncertain. Additionally, there could be potential risks with the data. There could be data privacy concerns due to the de-identified but still potentially identifiable patient information, potential biases in the data. We acknowledge the potential for bias when using LLMs as judges. To mitigate this risk, we incorporated several safeguards;
(a) LLM predictions are bench-marked against the human-annotated subset to ensure alignment and identify areas where LLMs might deviate (b) statistical power calculations guided the size of the human-labeled subset, ensuring it was large enough to rigorously validate LLM performance (c) our study explicitly discusses the limitations of LLM-based evaluation and highlights the importance of human-annotated benchmarks for future work.

In conclusion, we believe there is significant (and growing) utility to releasing large-scale synthetic and partially synthetic datasets in conjunction with LLM-based evaluators, reflected here by our \datasetname dataset and NLI clinician-curated dataset. This is especially important for academic labs that lack access to frontier model APIs that are HIPAA-compliant, creating barriers to exploring clinical text problems. Curating a sample of important clinical document types across multiple institutions and providing frontier LLM fact decompositions directly enables the research community to explore key problems in synthetic data quality evaluation, preference alignment, model distillation and more. By combining human-verified annotations with scalable LLM-based evaluations, our approach achieves a balance between scalability and the rigor needed for clinical applications, also keeping LLM bias in check \citep{li2024llmasajudge}. This also ensures reliability as we rigorously validated the performance of LLMs by benchmarking them against a carefully selected subset of manually labeled examples. We also explicitly discusses the limitations of LLM-based evaluation earlier in this section and highlights the importance of human-annotated benchmarks for future work.


\paragraph{Acknowledgments} 
This work is generously supported by the Mark and Debra Leslie endowment for AI in Healthcare. 
The authors also thank the Clinical Excellence Research Center (CERC) at Stanford for their support. 
AS thanks the Paul and Daisy Soros Foundation and Knight Hennessy Scholars for their support.

\bibliography{sample}

\begin{thebibliography}{54}
\providecommand{\natexlab}[1]{#1}
\providecommand{\url}[1]{\texttt{#1}}
\expandafter\ifx\csname urlstyle\endcsname\relax
  \providecommand{\doi}[1]{doi: #1}\else
  \providecommand{\doi}{doi: \begingroup \urlstyle{rm}\Url}\fi

\bibitem[Agrawal et~al.(2022)Agrawal, Hegselmann, Lang, Kim, and Sontag]{agrawal2022large}
Monica Agrawal, Stefan Hegselmann, Hunter Lang, Yoon Kim, and David Sontag.
\newblock Large language models are few-shot clinical information extractors, 2022.

\bibitem[Alsentzer et~al.(2019)Alsentzer, Murphy, Boag, Weng, Jindi, Naumann, and McDermott]{alsentzer-etal-2019-publicly}
Emily Alsentzer, John Murphy, William Boag, Wei-Hung Weng, Di~Jindi, Tristan Naumann, and Matthew McDermott.
\newblock Publicly available clinical {BERT} embeddings.
\newblock In Anna Rumshisky, Kirk Roberts, Steven Bethard, and Tristan Naumann, editors, \emph{Proceedings of the 2nd Clinical Natural Language Processing Workshop}, pages 72--78, Minneapolis, Minnesota, USA, June 2019. Association for Computational Linguistics.
\newblock \doi{10.18653/v1/W19-1909}.
\newblock URL \url{https://aclanthology.org/W19-1909}.

\bibitem[Arsenyan et~al.(2023)Arsenyan, Bughdaryan, Shaya, Small, and Shahnazaryan]{arsenyan2023llm}
Vahan Arsenyan, Spartak Bughdaryan, Fadi Shaya, Kent Small, and Davit Shahnazaryan.
\newblock Large language models for biomedical knowledge graph construction: Information extraction from emr notes, 2023.
\newblock URL \url{https://arxiv.org/abs/2301.12473}.
\newblock v2.

\bibitem[Bannur et~al.(2024)Bannur, Bouzid, Castro, Schwaighofer, Thieme, Bond-Taylor, Ilse, Pérez-García, Salvatelli, Sharma, Meissen, Ranjit, Srivastav, Gong, Codella, Falck, Oktay, Lungren, Wetscherek, Alvarez-Valle, and Hyland]{bannur2024maira2groundedradiologyreport}
Shruthi Bannur, Kenza Bouzid, Daniel~C. Castro, Anton Schwaighofer, Anja Thieme, Sam Bond-Taylor, Maximilian Ilse, Fernando Pérez-García, Valentina Salvatelli, Harshita Sharma, Felix Meissen, Mercy Ranjit, Shaury Srivastav, Julia Gong, Noel C.~F. Codella, Fabian Falck, Ozan Oktay, Matthew~P. Lungren, Maria~Teodora Wetscherek, Javier Alvarez-Valle, and Stephanie~L. Hyland.
\newblock Maira-2: Grounded radiology report generation, 2024.
\newblock URL \url{https://arxiv.org/abs/2406.04449}.

\bibitem[Bowman et~al.(2015)Bowman, Angeli, Potts, and Manning]{bowman-etal-2015-large}
Samuel~R. Bowman, Gabor Angeli, Christopher Potts, and Christopher~D. Manning.
\newblock A large annotated corpus for learning natural language inference.
\newblock In Llu{\'\i}s M{\`a}rquez, Chris Callison-Burch, and Jian Su, editors, \emph{Proceedings of the 2015 Conference on Empirical Methods in Natural Language Processing}, pages 632--642, Lisbon, Portugal, September 2015. Association for Computational Linguistics.
\newblock \doi{10.18653/v1/D15-1075}.
\newblock URL \url{https://aclanthology.org/D15-1075}.

\bibitem[Buderer(1996)]{https://doi.org/10.1111/j.1553-2712.1996.tb03538.x}
Nancy M.~Fenn Buderer.
\newblock Statistical methodology: I. incorporating the prevalence of disease into the sample size calculation for sensitivity and specificity.
\newblock \emph{Academic Emergency Medicine}, 3\penalty0 (9):\penalty0 895--900, 1996.
\newblock \doi{https://doi.org/10.1111/j.1553-2712.1996.tb03538.x}.
\newblock URL \url{https://onlinelibrary.wiley.com/doi/abs/10.1111/j.1553-2712.1996.tb03538.x}.

\bibitem[Ceballos-Arroyo et~al.(2024)Ceballos-Arroyo, Munnangi, Sun, Zhang, McInerney, Wallace, and Amir]{ceballos-arroyo-etal-2024-open}
Alberto~Mario Ceballos-Arroyo, Monica Munnangi, Jiuding Sun, Karen Zhang, Jered McInerney, Byron~C. Wallace, and Silvio Amir.
\newblock Open (clinical) {LLM}s are sensitive to instruction phrasings.
\newblock In Dina Demner-Fushman, Sophia Ananiadou, Makoto Miwa, Kirk Roberts, and Junichi Tsujii, editors, \emph{Proceedings of the 23rd Workshop on Biomedical Natural Language Processing}, pages 50--71, Bangkok, Thailand, August 2024. Association for Computational Linguistics.
\newblock \doi{10.18653/v1/2024.bionlp-1.5}.
\newblock URL \url{https://aclanthology.org/2024.bionlp-1.5/}.

\bibitem[Chang et~al.(2023)Chang, Wang, Wang, Wu, Yang, Zhu, Chen, Yi, Wang, Wang, Ye, Zhang, Chang, Yu, Yang, and Xie]{chang2023surveyevaluationlargelanguage}
Yupeng Chang, Xu~Wang, Jindong Wang, Yuan Wu, Linyi Yang, Kaijie Zhu, Hao Chen, Xiaoyuan Yi, Cunxiang Wang, Yidong Wang, Wei Ye, Yue Zhang, Yi~Chang, Philip~S. Yu, Qiang Yang, and Xing Xie.
\newblock A survey on evaluation of large language models, 2023.
\newblock URL \url{https://arxiv.org/abs/2307.03109}.

\bibitem[Chen et~al.(2023)Chen, Buthpitiya, Fabrikant, Roth, and Schuster]{chen-etal-2023-propsegment}
Sihao Chen, Senaka Buthpitiya, Alex Fabrikant, Dan Roth, and Tal Schuster.
\newblock {P}rop{S}egm{E}nt: A large-scale corpus for proposition-level segmentation and entailment recognition.
\newblock In Anna Rogers, Jordan Boyd-Graber, and Naoaki Okazaki, editors, \emph{Findings of the Association for Computational Linguistics: ACL 2023}, pages 8874--8893, Toronto, Canada, July 2023. Association for Computational Linguistics.
\newblock \doi{10.18653/v1/2023.findings-acl.565}.
\newblock URL \url{https://aclanthology.org/2023.findings-acl.565}.

\bibitem[Devaraj et~al.(2021)Devaraj, Marshall, Wallace, and Li]{devaraj-etal-2021-paragraph}
Ashwin Devaraj, Iain Marshall, Byron Wallace, and Junyi~Jessy Li.
\newblock Paragraph-level simplification of medical texts.
\newblock In Kristina Toutanova, Anna Rumshisky, Luke Zettlemoyer, Dilek Hakkani-Tur, Iz~Beltagy, Steven Bethard, Ryan Cotterell, Tanmoy Chakraborty, and Yichao Zhou, editors, \emph{Proceedings of the 2021 Conference of the North American Chapter of the Association for Computational Linguistics: Human Language Technologies}, pages 4972--4984, Online, June 2021. Association for Computational Linguistics.
\newblock \doi{10.18653/v1/2021.naacl-main.395}.
\newblock URL \url{https://aclanthology.org/2021.naacl-main.395}.

\bibitem[Dubey et~al.(2024)Dubey, Jauhri, Pandey, Kadian, Al-Dahle, Letman, Mathur, Schelten, Yang, and Angela~Fan]{grattafiori2024llama3herdmodels}
Abhimanyu Dubey, Abhinav Jauhri, Abhinav Pandey, Abhishek Kadian, Ahmad Al-Dahle, Aiesha Letman, Akhil Mathur, Alan Schelten, Amy Yang, and et~al. Angela~Fan.
\newblock The llama 3 herd of models, 2024.
\newblock URL \url{https://arxiv.org/abs/2407.21783}.

\bibitem[Everitt(2006)]{everitt2006cambridge}
Brian~S Everitt.
\newblock \emph{The Cambridge dictionary of statistics}.
\newblock Cambridge University Press 1998, 2002, 2006, 2006.

\bibitem[Fabbri et~al.(2022)Fabbri, Wu, Liu, and Xiong]{fabbri-etal-2022-qafacteval}
Alexander Fabbri, Chien-Sheng Wu, Wenhao Liu, and Caiming Xiong.
\newblock {QAF}act{E}val: Improved {QA}-based factual consistency evaluation for summarization.
\newblock In Marine Carpuat, Marie-Catherine de~Marneffe, and Ivan~Vladimir Meza~Ruiz, editors, \emph{Proceedings of the 2022 Conference of the North American Chapter of the Association for Computational Linguistics: Human Language Technologies}, pages 2587--2601, Seattle, United States, July 2022. Association for Computational Linguistics.
\newblock \doi{10.18653/v1/2022.naacl-main.187}.
\newblock URL \url{https://aclanthology.org/2022.naacl-main.187}.

\bibitem[Fleming et~al.(2023)Fleming, Lozano, Haberkorn, Jindal, Reis, Thapa, Blankemeier, Genkins, Steinberg, Nayak, Patel, Chiang, Callahan, Huo, Gatidis, Adams, Fayanju, Shah, Savage, Goh, Chaudhari, Aghaeepour, Sharp, Pfeffer, Liang, Chen, Morse, Brunskill, Fries, and Shah]{fleming2023medalign}
Scott~L. Fleming, Alejandro Lozano, William~J. Haberkorn, Jenelle~A. Jindal, Eduardo~P. Reis, Rahul Thapa, Louis Blankemeier, Julian~Z. Genkins, Ethan Steinberg, Ashwin Nayak, Birju~S. Patel, Chia-Chun Chiang, Alison Callahan, Zepeng Huo, Sergios Gatidis, Scott~J. Adams, Oluseyi Fayanju, Shreya~J. Shah, Thomas Savage, Ethan Goh, Akshay~S. Chaudhari, Nima Aghaeepour, Christopher Sharp, Michael~A. Pfeffer, Percy Liang, Jonathan~H. Chen, Keith~E. Morse, Emma~P. Brunskill, Jason~A. Fries, and Nigam~H. Shah.
\newblock Medalign: A clinician-generated dataset for instruction following with electronic medical records, 2023.

\bibitem[Gunjal and Durrett(2024)]{gunjal2024molecularfactsdesideratadecontextualization}
Anisha Gunjal and Greg Durrett.
\newblock Molecular facts: Desiderata for decontextualization in llm fact verification, 2024.
\newblock URL \url{https://arxiv.org/abs/2406.20079}.

\bibitem[Hegselmann et~al.(2024)Hegselmann, Shen, Gierse, Agrawal, Sontag, and Jiang]{hegselmann2024datacentricapproachgeneratefaithful}
Stefan Hegselmann, Shannon~Zejiang Shen, Florian Gierse, Monica Agrawal, David Sontag, and Xiaoyi Jiang.
\newblock A data-centric approach to generate faithful and high quality patient summaries with large language models, 2024.
\newblock URL \url{https://arxiv.org/abs/2402.15422}.

\bibitem[Jain et~al.(2021)Jain, Agrawal, Saporta, Truong, Duong, Bui, Chambon, Zhang, Lungren, Ng, Langlotz, and Rajpurkar]{jain2021radgraphextractingclinicalentities}
Saahil Jain, Ashwin Agrawal, Adriel Saporta, Steven~QH Truong, Du~Nguyen Duong, Tan Bui, Pierre Chambon, Yuhao Zhang, Matthew~P. Lungren, Andrew~Y. Ng, Curtis~P. Langlotz, and Pranav Rajpurkar.
\newblock Radgraph: Extracting clinical entities and relations from radiology reports, 2021.
\newblock URL \url{https://arxiv.org/abs/2106.14463}.

\bibitem[Jeong et~al.(2024)Jeong, Garg, Lipton, and Oberst]{jeong2024medicaladaptationlargelanguage}
Daniel~P. Jeong, Saurabh Garg, Zachary~C. Lipton, and Michael Oberst.
\newblock Medical adaptation of large language and vision-language models: Are we making progress?, 2024.
\newblock URL \url{https://arxiv.org/abs/2411.04118}.

\bibitem[Johnson et~al.(2016)Johnson, Pollard, Shen, wei H.~Lehman, Feng, Ghassemi, Moody, Szolovits, Celi, and Mark]{Johnson2016MIMICIIIAF}
Alistair E.~W. Johnson, Tom~J. Pollard, Lu~Shen, Li~wei H.~Lehman, Mengling Feng, Mohammad~Mahdi Ghassemi, Benjamin Moody, Peter Szolovits, Leo~Anthony Celi, and Roger~G. Mark.
\newblock Mimic-iii, a freely accessible critical care database.
\newblock \emph{Scientific Data}, 3, 2016.
\newblock URL \url{https://api.semanticscholar.org/CorpusID:33285731}.

\bibitem[Johnson et~al.(2019)Johnson, Pollard, Berkowitz, Greenbaum, Lungren, Deng, Mark, and Horng]{johnson2019mimic}
Alistair~EW Johnson, Tom~J Pollard, Seth~J Berkowitz, Nathaniel~R Greenbaum, Matthew~P Lungren, Chih-ying Deng, Roger~G Mark, and Steven Horng.
\newblock Mimic-cxr, a de-identified publicly available database of chest radiographs with free-text reports.
\newblock \emph{Scientific data}, 6\penalty0 (1):\penalty0 317, 2019.

\bibitem[Jullien et~al.(2023)Jullien, Valentino, Frost, O{'}Regan, Landers, and Freitas]{jullien-etal-2023-nli4ct}
Mael Jullien, Marco Valentino, Hannah Frost, Paul O{'}Regan, D{\'o}nal Landers, and Andre Freitas.
\newblock {NLI}4{CT}: Multi-evidence natural language inference for clinical trial reports.
\newblock In Houda Bouamor, Juan Pino, and Kalika Bali, editors, \emph{Proceedings of the 2023 Conference on Empirical Methods in Natural Language Processing}, pages 16745--16764, Singapore, December 2023. Association for Computational Linguistics.
\newblock \doi{10.18653/v1/2023.emnlp-main.1041}.
\newblock URL \url{https://aclanthology.org/2023.emnlp-main.1041/}.

\bibitem[Kamoi et~al.(2023)Kamoi, Goyal, Diego~Rodriguez, and Durrett]{kamoi-etal-2023-wice}
Ryo Kamoi, Tanya Goyal, Juan Diego~Rodriguez, and Greg Durrett.
\newblock {W}i{CE}: Real-world entailment for claims in {W}ikipedia.
\newblock In Houda Bouamor, Juan Pino, and Kalika Bali, editors, \emph{Proceedings of the 2023 Conference on Empirical Methods in Natural Language Processing}, pages 7561--7583, Singapore, December 2023. Association for Computational Linguistics.
\newblock \doi{10.18653/v1/2023.emnlp-main.470}.
\newblock URL \url{https://aclanthology.org/2023.emnlp-main.470}.

\bibitem[Khot et~al.(2018)Khot, Sabharwal, and Clark]{DBLP:conf/aaai/KhotSC18}
Tushar Khot, Ashish Sabharwal, and Peter Clark.
\newblock Scitail: {A} textual entailment dataset from science question answering.
\newblock In Sheila~A. McIlraith and Kilian~Q. Weinberger, editors, \emph{Proceedings of the Thirty-Second {AAAI} Conference on Artificial Intelligence, (AAAI-18), the 30th innovative Applications of Artificial Intelligence (IAAI-18), and the 8th {AAAI} Symposium on Educational Advances in Artificial Intelligence (EAAI-18), New Orleans, Louisiana, USA, February 2-7, 2018}, pages 5189--5197. {AAAI} Press, 2018.
\newblock \doi{10.1609/AAAI.V32I1.12022}.
\newblock URL \url{https://doi.org/10.1609/aaai.v32i1.12022}.

\bibitem[Kusner et~al.(2015)Kusner, Sun, Kolkin, and Weinberger]{kusner2015word}
Matt Kusner, Yu~Sun, Nicholas Kolkin, and Kilian Weinberger.
\newblock From word embeddings to document distances.
\newblock In \emph{International conference on machine learning}, pages 957--966. PMLR, 2015.

\bibitem[Lelkes et~al.(2023)Lelkes, Loreaux, Schuster, Chen, and Rajkomar]{lelkes-etal-2023-sdoh}
Adam Lelkes, Eric Loreaux, Tal Schuster, Ming-Jun Chen, and Alvin Rajkomar.
\newblock {SDOH}-{NLI}: a dataset for inferring social determinants of health from clinical notes.
\newblock In Houda Bouamor, Juan Pino, and Kalika Bali, editors, \emph{Findings of the Association for Computational Linguistics: EMNLP 2023}, pages 4789--4798, Singapore, December 2023. Association for Computational Linguistics.
\newblock \doi{10.18653/v1/2023.findings-emnlp.317}.
\newblock URL \url{https://aclanthology.org/2023.findings-emnlp.317/}.

\bibitem[Li et~al.(2024)Li, Jiang, Huang, Beigi, Zhao, Tan, Bhattacharjee, Jiang, Chen, Wu, Shu, Cheng, and Liu]{li2024llmasajudge}
Dawei Li, Bohan Jiang, Liangjie Huang, Alimohammad Beigi, Chengshuai Zhao, Zhen Tan, Amrita Bhattacharjee, Yuxuan Jiang, Canyu Chen, Tianhao Wu, Kai Shu, Lu~Cheng, and Huan Liu.
\newblock From generation to judgment: Opportunities and challenges of llm-as-a-judge.
\newblock \emph{arXiv preprint arXiv: 2411.16594}, 2024.

\bibitem[Lin(2004)]{lin-2004-rouge}
Chin-Yew Lin.
\newblock {ROUGE}: A package for automatic evaluation of summaries.
\newblock In \emph{Text Summarization Branches Out}, pages 74--81, Barcelona, Spain, July 2004. Association for Computational Linguistics.
\newblock URL \url{https://aclanthology.org/W04-1013/}.

\bibitem[Min et~al.(2023)Min, Krishna, Lyu, Lewis, tau Yih, Koh, Iyyer, Zettlemoyer, and Hajishirzi]{min2023factscore}
Sewon Min, Kalpesh Krishna, Xinxi Lyu, Mike Lewis, Wen tau Yih, Pang~Wei Koh, Mohit Iyyer, Luke Zettlemoyer, and Hannaneh Hajishirzi.
\newblock Factscore: Fine-grained atomic evaluation of factual precision in long form text generation, 2023.

\bibitem[Miura et~al.(2021)Miura, Zhang, Tsai, Langlotz, and Jurafsky]{miura-etal-2021-improving}
Yasuhide Miura, Yuhao Zhang, Emily Tsai, Curtis Langlotz, and Dan Jurafsky.
\newblock Improving factual completeness and consistency of image-to-text radiology report generation.
\newblock In Kristina Toutanova, Anna Rumshisky, Luke Zettlemoyer, Dilek Hakkani-Tur, Iz~Beltagy, Steven Bethard, Ryan Cotterell, Tanmoy Chakraborty, and Yichao Zhou, editors, \emph{Proceedings of the 2021 Conference of the North American Chapter of the Association for Computational Linguistics: Human Language Technologies}, pages 5288--5304, Online, June 2021. Association for Computational Linguistics.
\newblock \doi{10.18653/v1/2021.naacl-main.416}.
\newblock URL \url{https://aclanthology.org/2021.naacl-main.416}.

\bibitem[Munnangi et~al.(2024)Munnangi, Feldman, Wallace, Amir, Hope, and Naik]{munnangi-etal-2024-fly}
Monica Munnangi, Sergey Feldman, Byron Wallace, Silvio Amir, Tom Hope, and Aakanksha Naik.
\newblock On-the-fly definition augmentation of {LLM}s for biomedical {NER}.
\newblock In Kevin Duh, Helena Gomez, and Steven Bethard, editors, \emph{Proceedings of the 2024 Conference of the North American Chapter of the Association for Computational Linguistics: Human Language Technologies (Volume 1: Long Papers)}, pages 3833--3854, Mexico City, Mexico, June 2024. Association for Computational Linguistics.
\newblock \doi{10.18653/v1/2024.naacl-long.212}.
\newblock URL \url{https://aclanthology.org/2024.naacl-long.212/}.

\bibitem[OpenAI(2023)]{openai2023gpt4}
OpenAI.
\newblock Gpt-4 technical report, 2023.

\bibitem[OpenAI(2024)]{openai2024gpt4ocard}
OpenAI.
\newblock Gpt-4o system card, 2024.
\newblock URL \url{https://arxiv.org/abs/2410.21276}.

\bibitem[Papineni et~al.(2002)Papineni, Roukos, Ward, and Zhu]{papineni-etal-2002-bleu}
Kishore Papineni, Salim Roukos, Todd Ward, and Wei-Jing Zhu.
\newblock {B}leu: a method for automatic evaluation of machine translation.
\newblock In Pierre Isabelle, Eugene Charniak, and Dekang Lin, editors, \emph{Proceedings of the 40th Annual Meeting of the Association for Computational Linguistics}, pages 311--318, Philadelphia, Pennsylvania, USA, July 2002. Association for Computational Linguistics.
\newblock \doi{10.3115/1073083.1073135}.
\newblock URL \url{https://aclanthology.org/P02-1040/}.

\bibitem[Radhakrishnan et~al.(2023)Radhakrishnan, Schenk, Ferar, Oskotsky, Ashouri~Choshali, Plunkett, Israni, and Butte]{article}
Lakshmi Radhakrishnan, Gundolf Schenk, Kathleen Ferar, Boris Oskotsky, Habibeh Ashouri~Choshali, Thomas Plunkett, Sharat Israni, and Atul Butte.
\newblock A certified de-identification system for all clinical text documents for information extraction at scale.
\newblock \emph{JAMIA Open}, 6, 07 2023.
\newblock \doi{10.1093/jamiaopen/ooad045}.

\bibitem[Romanov and Shivade(2018{\natexlab{a}})]{romanov-shivade-2018-lessons}
Alexey Romanov and Chaitanya Shivade.
\newblock Lessons from natural language inference in the clinical domain.
\newblock In Ellen Riloff, David Chiang, Julia Hockenmaier, and Jun{'}ichi Tsujii, editors, \emph{Proceedings of the 2018 Conference on Empirical Methods in Natural Language Processing}, pages 1586--1596, Brussels, Belgium, October-November 2018{\natexlab{a}}. Association for Computational Linguistics.
\newblock \doi{10.18653/v1/D18-1187}.
\newblock URL \url{https://aclanthology.org/D18-1187}.

\bibitem[Romanov and Shivade(2018{\natexlab{b}})]{romanov2018lessonsnaturallanguageinference}
Alexey Romanov and Chaitanya Shivade.
\newblock Lessons from natural language inference in the clinical domain, 2018{\natexlab{b}}.
\newblock URL \url{https://arxiv.org/abs/1808.06752}.

\bibitem[Ru et~al.(2024)Ru, Qiu, Hu, Zhang, Shi, Chang, Cheng, Wang, Sun, Li, Zhang, Wang, Jiang, He, Wang, Liu, Zhang, and Zhang]{ru2024ragcheckerfinegrainedframeworkdiagnosing}
Dongyu Ru, Lin Qiu, Xiangkun Hu, Tianhang Zhang, Peng Shi, Shuaichen Chang, Jiayang Cheng, Cunxiang Wang, Shichao Sun, Huanyu Li, Zizhao Zhang, Binjie Wang, Jiarong Jiang, Tong He, Zhiguo Wang, Pengfei Liu, Yue Zhang, and Zheng Zhang.
\newblock Ragchecker: A fine-grained framework for diagnosing retrieval-augmented generation, 2024.
\newblock URL \url{https://arxiv.org/abs/2408.08067}.

\bibitem[Rubner et~al.(1998)Rubner, Tomasi, and Guibas]{rubner1998metric}
Yossi Rubner, Carlo Tomasi, and Leonidas~J Guibas.
\newblock A metric for distributions with applications to image databases.
\newblock In \emph{Sixth international conference on computer vision (IEEE Cat. No. 98CH36271)}, pages 59--66. IEEE, 1998.

\bibitem[Singhal et~al.(2023{\natexlab{a}})Singhal, Azizi, Tu, Mahdavi, Wei, Chung, Scales, Tanwani, Cole-Lewis, Pfohl, et~al.]{singhal2023large}
Karan Singhal, Shekoofeh Azizi, Tao Tu, S~Sara Mahdavi, Jason Wei, Hyung~Won Chung, Nathan Scales, Ajay Tanwani, Heather Cole-Lewis, Stephen Pfohl, et~al.
\newblock Large language models encode clinical knowledge.
\newblock \emph{Nature}, 620\penalty0 (7972):\penalty0 172--180, 2023{\natexlab{a}}.

\bibitem[Singhal et~al.(2023{\natexlab{b}})Singhal, Tu, Gottweis, Sayres, Wulczyn, Hou, Clark, Pfohl, Cole-Lewis, Neal, Schaekermann, Wang, Amin, Lachgar, Mansfield, Prakash, Green, Dominowska, y~Arcas, Tomasev, Liu, Wong, Semturs, Mahdavi, Barral, Webster, Corrado, Matias, Azizi, Karthikesalingam, and Natarajan]{singhal2023expertlevelmedicalquestionanswering}
Karan Singhal, Tao Tu, Juraj Gottweis, Rory Sayres, Ellery Wulczyn, Le~Hou, Kevin Clark, Stephen Pfohl, Heather Cole-Lewis, Darlene Neal, Mike Schaekermann, Amy Wang, Mohamed Amin, Sami Lachgar, Philip Mansfield, Sushant Prakash, Bradley Green, Ewa Dominowska, Blaise~Aguera y~Arcas, Nenad Tomasev, Yun Liu, Renee Wong, Christopher Semturs, S.~Sara Mahdavi, Joelle Barral, Dale Webster, Greg~S. Corrado, Yossi Matias, Shekoofeh Azizi, Alan Karthikesalingam, and Vivek Natarajan.
\newblock Towards expert-level medical question answering with large language models, 2023{\natexlab{b}}.
\newblock URL \url{https://arxiv.org/abs/2305.09617}.

\bibitem[Sushil et~al.(2024)Sushil, Kennedy, Mandair, Miao, Zack, and Butte]{doi:10.1056/AIdbp2300110}
Madhumita Sushil, Vanessa~E. Kennedy, Divneet Mandair, Brenda~Y. Miao, Travis Zack, and Atul~J. Butte.
\newblock Coral: Expert-curated oncology reports to advance language model inference.
\newblock \emph{NEJM AI}, 0\penalty0 (0):\penalty0 AIdbp2300110, 2024.
\newblock \doi{10.1056/AIdbp2300110}.
\newblock URL \url{https://ai.nejm.org/doi/abs/10.1056/AIdbp2300110}.

\bibitem[Team et~al.(2024)Team, Georgiev, Lei, Burnell, Bai, Gulati, Tanzer, Vincent, Pan, Wang, Mariooryad, Ding, Geng, Alcober, Frostig, Omernick, Walker, Paduraru, Sorokin, Tacchetti, Gaffney, Daruki, Sercinoglu, Gleicher, Love, Voigtlaender, Jain, Surita, Mohamed, Blevins, Ahn, Zhu, Kawintiranon, Firat, Gu, Zhang, Rahtz, Faruqui, Clay, Gilmer, Co-Reyes, Penchev, Zhu, Morioka, Hui, Haridasan, Campos, Mahdieh, Guo, Hassan, Kilgour, Vezer, Cheng, de~Liedekerke, Goyal, Barham, Strouse, Noury, Adler, Sundararajan, Vikram, Lepikhin, Paganini, Garcia, Yang, Valter, Trebacz, Vodrahalli, Asawaroengchai, Ring, Kalb, Soares, Brahma, Steiner, Yu, Mentzer, He, Gonzalez, Xu, Kaufman, Shafey, Oh, Hennigan, van~den Driessche, Odoom, Lucic, Roelofs, Lall, Marathe, Chan, Ontanon, He, Teplyashin, Lai, Crone, Damoc, Ho, Riedel, Lenc, Yeh, Chowdhery, Xu, Kazemi, Amid, Petrushkina, Swersky, Khodaei, Chen, Larkin, Pinto, Yan, Badia, Patil, Hansen, Orr, Arnold, Grimstad, Dai, Douglas, Sinha, Yadav, Chen, Gribovskaya, Austin,
  Zhao, Patel, Komarek, Austin, Borgeaud, Friso, Goyal, Caine, Cao, Chung, Lamm, Barth-Maron, Kagohara, Olszewska, Chen, Shivakumar, Agarwal, Godhia, Rajwar, Snaider, Dotiwalla, Liu, Barua, Ungureanu, Zhang, Batsaikhan, Wirth, Qin, Danihelka, Doshi, Chadwick, Chen, Jain, Le, Kar, Gurumurthy, Li, Sang, Liu, Lamprou, Munoz, Lintz, Mehta, Howard, Reynolds, Aroyo, Wang, Blanco, Cassirer, Griffith, Das, Lee, Sygnowski, Fisher, Besley, Powell, Ahmed, Paulus, Reitter, Borsos, Joshi, Pope, Hand, Selo, Jain, Sethi, Goel, Makino, May, Yang, Schalkwyk, Butterfield, Hauth, Goldin, Hawkins, Senter, Brin, Woodman, Ritter, Noland, Giang, Bolina, Lee, Blyth, Mackinnon, Reid, Sarvana, Silver, Chen, Wang, Maggiore, Chang, Attaluri, Thornton, Chiu, Bunyan, Levine, Chung, Eltyshev, Si, Lillicrap, Brady, Aggarwal, Wu, Xu, McIlroy, Badola, Sandhu, Moreira, Stokowiec, Hemsley, Li, Tudor, Shyam, Rahimtoroghi, Haykal, Sprechmann, Zhou, Mincu, Li, Addanki, Krishna, Wu, Frechette, Eyal, Dafoe, Lacey, Whang, Avrahami, Zhang, Taropa,
  Lin, Toyama, Rutherford, Sano, Choe, Tomala, Safranek-Shrader, Kassner, Pajarskas, Harvey, Sechrist, Fortunato, Lyu, Elsayed, Kuang, Lottes, Chu, Jia, Chen, Humphreys, Baumli, Tao, Samuel, dos Santos, Andreassen, Rakićević, Grewe, Kumar, Winkler, Caton, Brock, Dalmia, Sheahan, Barr, Miao, Natsev, Devlin, Behbahani, Prost, Sun, Myaskovsky, Pillai, Hurt, Lazaridou, Xiong, Zheng, Pardo, Li, Horgan, Stanton, Ambar, Xia, Lince, Wang, Mustafa, Webson, Lee, Anil, Wicke, Dozat, Sinha, Piqueras, Dabir, Upadhyay, Boral, Hendricks, Fry, Djolonga, Su, Walker, Labanowski, Huang, Misra, Chen, Skerry-Ryan, Singh, Rijhwani, Yu, Castro-Ros, Changpinyo, Datta, Bagri, Hrafnkelsson, Maggioni, Zheng, Sulsky, Hou, Paine, Yang, Riesa, Rogozinska, Marcus, Badawy, Zhang, Wang, Miller, Greer, Sjos, Nova, Zen, Chaabouni, Rosca, Jiang, Chen, Liu, Sainath, Krikun, Polozov, Lespiau, Newlan, Cankara, Kwak, Xu, Chen, Coenen, Meyer, Tsihlas, Ma, Gottweis, Xing, Gu, Miao, Frank, Cankara, Ganapathy, Dasgupta, Hughes-Fitt, Chen, Reid, Rong,
  Fan, van Amersfoort, Zhuang, Cohen, Gu, Mohananey, Ilic, Tobin, Wieting, Bortsova, Thacker, Wang, Caveness, Chiu, Sezener, Kaskasoli, Baker, Millican, Elhawaty, Aisopos, Lebsack, Byrd, Dai, Jia, Wiethoff, Davoodi, Weston, Yagati, Ahuja, Gao, Pundak, Zhang, Azzam, Sim, Caelles, Keeling, Sharma, Swing, Li, Liu, Bostock, Bansal, Nado, Anand, Lipschultz, Karmarkar, Proleev, Ittycheriah, Yeganeh, Polovets, Faust, Sun, Rrustemi, Li, Shivanna, Liu, Welty, Lebron, Baddepudi, Krause, Parisotto, Soricut, Xu, Bloxwich, Johnson, Neyshabur, Mao-Jones, Wang, Ramasesh, Abbas, Guez, Segal, Nguyen, Svensson, Hou, York, Milan, Bridgers, Gworek, Tagliasacchi, Lee-Thorp, Chang, Guseynov, Hartman, Kwong, Zhao, Kashem, Cole, Miech, Tanburn, Phuong, Pavetic, Cevey, Comanescu, Ives, Yang, Du, Li, Zhang, Iinuma, Hu, Roy, Bijwadia, Zhu, Martins, Saputro, Gergely, Zheng, Jia, Antonoglou, Sadovsky, Gu, Bi, Andreev, Samangooei, Khan, Kocisky, Filos, Kumar, Bishop, Yu, Hodkinson, Mittal, Shah, Moufarek, Cheng, Bloniarz, Lee, Pejman,
  Michel, Spencer, Feinberg, Xiong, Savinov, Smith, Shakeri, Tran, Chesus, Bohnet, Tucker, von Glehn, Muir, Mao, Kazawa, Slone, Soparkar, Shrivastava, Cobon-Kerr, Sharman, Pavagadhi, Araya, Misiunas, Ghelani, Laskin, Barker, Li, Briukhov, Houlsby, Glaese, Lakshminarayanan, Schucher, Tang, Collins, Lim, Feng, Recasens, Lai, Magni, Cao, Siddhant, Ashwood, Orbay, Dehghani, Brennan, He, Xu, Gao, Saroufim, Molloy, Wu, Arnold, Chang, Schrittwieser, Buchatskaya, Radpour, Polacek, Giordano, Bapna, Tokumine, Hellendoorn, Sottiaux, Cogan, Severyn, Saleh, Thakoor, Shefey, Qiao, Gaba, yiin Chang, Swanson, Zhang, Lee, Rubenstein, Song, Kwiatkowski, Koop, Kannan, Kao, Schuh, Stjerngren, Ghiasi, Gibson, Vilnis, Yuan, Ferreira, Kamath, Klimenko, Franko, Xiao, Bhattacharya, Patel, Wang, Morris, Strudel, Sharma, Choy, Hashemi, Landon, Finkelstein, Jhakra, Frye, Barnes, Mauger, Daun, Baatarsukh, Tung, Farhan, Michalewski, Viola, de~Chaumont~Quitry, Lan, Hudson, Wang, Fischer, Zheng, White, Dragan, baptiste Alayrac, Ni, Pritzel,
  Iwanicki, Isard, Bulanova, Zilka, Dyer, Sachan, Srinivasan, Muckenhirn, Cai, Mandhane, Tariq, Rae, Wang, Ayoub, FitzGerald, Zhao, Han, Alberti, Garrette, Krishnakumar, Gimenez, Levskaya, Sohn, Matak, Iturrate, Chang, Xiang, Cao, Ranka, Brown, Hutter, Mirrokni, Chen, Yao, Egyed, Galilee, Liechty, Kallakuri, Palmer, Ghemawat, Liu, Tao, Thornton, Green, Jasarevic, Lin, Cotruta, Tan, Fiedel, Yu, Chi, Neitz, Heitkaemper, Sinha, Zhou, Sun, Kaed, Hulse, Mishra, Georgaki, Kudugunta, Farabet, Shafran, Vlasic, Tsitsulin, Ananthanarayanan, Carin, Su, Sun, V, Carvajal, Broder, Comsa, Repina, Wong, Chen, Hawkins, Filonov, Loher, Hirnschall, Wang, Ye, Burns, Cate, Wright, Piccinini, Zhang, Lin, Gog, Kulizhskaya, Sreevatsa, Song, Cobo, Iyer, Tekur, Garrido, Xiao, Kemp, Zheng, Li, Agarwal, Ngani, Goshvadi, Santamaria-Fernandez, Fica, Chen, Gorgolewski, Sun, Garg, Ye, Eslami, Hua, Simon, Joshi, Kim, Tenney, Potluri, Thiet, Yuan, Luisier, Chronopoulou, Scellato, Srinivasan, Chen, Koverkathu, Dalibard, Xu, Saeta, Anderson,
  Sellam, Fernando, Huot, Jung, Varadarajan, Quinn, Raul, Le, Habalov, Clark, Jalan, Bullard, Singhal, Luong, Wang, Rajayogam, Eisenschlos, Jia, Finchelstein, Yakubovich, Balle, Fink, Agarwal, Li, Dvijotham, Pal, Kang, Konzelmann, Beattie, Dousse, Wu, Crocker, Elkind, Jonnalagadda, Lee, Holtmann-Rice, Kallarackal, Liu, Vnukov, Vats, Invernizzi, Jafari, Zhou, Taylor, Prendki, Wu, Eccles, Liu, Kopparapu, Beaufays, Angermueller, Marzoca, Sarcar, Dib, Stanway, Perbet, Trdin, Sterneck, Khorlin, Li, Wu, Goenka, Madras, Goldshtein, Gierke, Zhou, Liu, Liang, White, Li, Singh, Bahargam, Epstein, Basu, Lao, Ozturel, Crous, Zhai, Lu, Tung, Gaur, Walton, Dixon, Zhang, Globerson, Uy, Bolt, Wiles, Nasr, Shumailov, Selvi, Piccinno, Aguilar, McCarthy, Khalman, Shukla, Galic, Carpenter, Villela, Zhang, Richardson, Martens, Bosnjak, Belle, Seibert, Alnahlawi, McWilliams, Singh, Louis, Ding, Popovici, Simicich, Knight, Mehta, Gupta, Shi, Fatehi, Mitrovic, Grills, Pagadora, Munkhdalai, Petrova, Eisenbud, Zhang, Yates, Mittal,
  Tripuraneni, Assael, Brovelli, Jain, Velimirovic, Akbulut, Mu, Macherey, Kumar, Xu, Qureshi, Comanici, Wiesner, Gong, Ruddock, Bauer, Felt, GP, Arnab, Zelle, Rothfuss, Rosgen, Shenoy, Seybold, Li, Mudigonda, Erdogan, Xia, Simsa, Michi, Yao, Yew, Kan, Caswell, Radebaugh, Elisseeff, Valenzuela, McKinney, Paterson, Cui, Latorre-Chimoto, Kim, Zeng, Durden, Ponnapalli, Sosea, Choquette-Choo, Manyika, Robenek, Vashisht, Pereira, Lam, Velic, Owusu-Afriyie, Lee, Bolukbasi, Parrish, Lu, Park, Venkatraman, Talbert, Rosique, Cheng, Sozanschi, Paszke, Kumar, Austin, Li, Salama, Perz, Kim, Dukkipati, Baryshnikov, Kaplanis, Sheng, Chervonyi, Unlu, de~Las~Casas, Askham, Tunyasuvunakool, Gimeno, Poder, Kwak, Miecnikowski, Mirrokni, Dimitriev, Parisi, Liu, Tsai, Shevlane, Kouridi, Garmon, Goedeckemeyer, Brown, Vijayakumar, Elqursh, Jazayeri, Huang, Carthy, Hoover, Kim, Kumar, Chen, Biles, Bingham, Rosen, Wang, Tan, Engel, Pongetti, de~Cesare, Hwang, Yu, Pullman, Narayanan, Levin, Gopal, Li, Aharoni, Trinh, Lo, Casagrande,
  Vij, Matthey, Ramadhana, Matthews, Carey, Johnson, Goranova, Shah, Ashraf, Dasgupta, Larsen, Wang, Vuyyuru, Jiang, Ijazi, Osawa, Smith, Boppana, Bilal, Koizumi, Xu, Altun, Shabat, Bariach, Korchemniy, Choo, Ronneberger, Iwuanyanwu, Zhao, Soergel, Hsieh, Cai, Iqbal, Sundermeyer, Chen, Bursztein, Malaviya, Biadsy, Shroff, Dhillon, Latkar, Dyer, Forbes, Nicosia, Nikolaev, Greene, Georgiev, Wang, Martin, Sedghi, Zhang, Banzal, Fritz, Rao, Wang, Zhang, Patraucean, Du, Mordatch, Jurin, Liu, Dubey, Mohan, Nowakowski, Ion, Wei, Tojo, Raad, Hudson, Keshava, Agrawal, Ramirez, Wu, Nguyen, Liu, Sewak, Petrini, Choi, Philips, Wang, Bica, Garg, Wilkiewicz, Agrawal, Li, Guo, Xue, Shaik, Leach, Khan, Wiesinger, Jerome, Chakladar, Wang, Ornduff, Abu, Ghaffarkhah, Wainwright, Cortes, Liu, Maynez, Terzis, Samangouei, Mansour, Kępa, Aubet, Algymr, Banica, Weisz, Orban, Senges, Andrejczuk, Geller, Santo, Anklin, Merey, Baeuml, Strohman, Bai, Petrov, Wu, Hassabis, Kavukcuoglu, Dean, and
  Vinyals]{geminiteam2024gemini15unlockingmultimodal}
Gemini Team, Petko Georgiev, Ving~Ian Lei, Ryan Burnell, Libin Bai, Anmol Gulati, Garrett Tanzer, Damien Vincent, Zhufeng Pan, Shibo Wang, Soroosh Mariooryad, Yifan Ding, Xinyang Geng, Fred Alcober, Roy Frostig, Mark Omernick, Lexi Walker, Cosmin Paduraru, Christina Sorokin, Andrea Tacchetti, Colin Gaffney, Samira Daruki, Olcan Sercinoglu, Zach Gleicher, Juliette Love, Paul Voigtlaender, Rohan Jain, Gabriela Surita, Kareem Mohamed, Rory Blevins, Junwhan Ahn, Tao Zhu, Kornraphop Kawintiranon, Orhan Firat, Yiming Gu, Yujing Zhang, Matthew Rahtz, Manaal Faruqui, Natalie Clay, Justin Gilmer, JD~Co-Reyes, Ivo Penchev, Rui Zhu, Nobuyuki Morioka, Kevin Hui, Krishna Haridasan, Victor Campos, Mahdis Mahdieh, Mandy Guo, Samer Hassan, Kevin Kilgour, Arpi Vezer, Heng-Tze Cheng, Raoul de~Liedekerke, Siddharth Goyal, Paul Barham, DJ~Strouse, Seb Noury, Jonas Adler, Mukund Sundararajan, Sharad Vikram, Dmitry Lepikhin, Michela Paganini, Xavier Garcia, Fan Yang, Dasha Valter, Maja Trebacz, Kiran Vodrahalli, Chulayuth
  Asawaroengchai, Roman Ring, Norbert Kalb, Livio~Baldini Soares, Siddhartha Brahma, David Steiner, Tianhe Yu, Fabian Mentzer, Antoine He, Lucas Gonzalez, Bibo Xu, Raphael~Lopez Kaufman, Laurent~El Shafey, Junhyuk Oh, Tom Hennigan, George van~den Driessche, Seth Odoom, Mario Lucic, Becca Roelofs, Sid Lall, Amit Marathe, Betty Chan, Santiago Ontanon, Luheng He, Denis Teplyashin, Jonathan Lai, Phil Crone, Bogdan Damoc, Lewis Ho, Sebastian Riedel, Karel Lenc, Chih-Kuan Yeh, Aakanksha Chowdhery, Yang Xu, Mehran Kazemi, Ehsan Amid, Anastasia Petrushkina, Kevin Swersky, Ali Khodaei, Gowoon Chen, Chris Larkin, Mario Pinto, Geng Yan, Adria~Puigdomenech Badia, Piyush Patil, Steven Hansen, Dave Orr, Sebastien M.~R. Arnold, Jordan Grimstad, Andrew Dai, Sholto Douglas, Rishika Sinha, Vikas Yadav, Xi~Chen, Elena Gribovskaya, Jacob Austin, Jeffrey Zhao, Kaushal Patel, Paul Komarek, Sophia Austin, Sebastian Borgeaud, Linda Friso, Abhimanyu Goyal, Ben Caine, Kris Cao, Da-Woon Chung, Matthew Lamm, Gabe Barth-Maron, Thais
  Kagohara, Kate Olszewska, Mia Chen, Kaushik Shivakumar, Rishabh Agarwal, Harshal Godhia, Ravi Rajwar, Javier Snaider, Xerxes Dotiwalla, Yuan Liu, Aditya Barua, Victor Ungureanu, Yuan Zhang, Bat-Orgil Batsaikhan, Mateo Wirth, James Qin, Ivo Danihelka, Tulsee Doshi, Martin Chadwick, Jilin Chen, Sanil Jain, Quoc Le, Arjun Kar, Madhu Gurumurthy, Cheng Li, Ruoxin Sang, Fangyu Liu, Lampros Lamprou, Rich Munoz, Nathan Lintz, Harsh Mehta, Heidi Howard, Malcolm Reynolds, Lora Aroyo, Quan Wang, Lorenzo Blanco, Albin Cassirer, Jordan Griffith, Dipanjan Das, Stephan Lee, Jakub Sygnowski, Zach Fisher, James Besley, Richard Powell, Zafarali Ahmed, Dominik Paulus, David Reitter, Zalan Borsos, Rishabh Joshi, Aedan Pope, Steven Hand, Vittorio Selo, Vihan Jain, Nikhil Sethi, Megha Goel, Takaki Makino, Rhys May, Zhen Yang, Johan Schalkwyk, Christina Butterfield, Anja Hauth, Alex Goldin, Will Hawkins, Evan Senter, Sergey Brin, Oliver Woodman, Marvin Ritter, Eric Noland, Minh Giang, Vijay Bolina, Lisa Lee, Tim Blyth, Ian
  Mackinnon, Machel Reid, Obaid Sarvana, David Silver, Alexander Chen, Lily Wang, Loren Maggiore, Oscar Chang, Nithya Attaluri, Gregory Thornton, Chung-Cheng Chiu, Oskar Bunyan, Nir Levine, Timothy Chung, Evgenii Eltyshev, Xiance Si, Timothy Lillicrap, Demetra Brady, Vaibhav Aggarwal, Boxi Wu, Yuanzhong Xu, Ross McIlroy, Kartikeya Badola, Paramjit Sandhu, Erica Moreira, Wojciech Stokowiec, Ross Hemsley, Dong Li, Alex Tudor, Pranav Shyam, Elahe Rahimtoroghi, Salem Haykal, Pablo Sprechmann, Xiang Zhou, Diana Mincu, Yujia Li, Ravi Addanki, Kalpesh Krishna, Xiao Wu, Alexandre Frechette, Matan Eyal, Allan Dafoe, Dave Lacey, Jay Whang, Thi Avrahami, Ye~Zhang, Emanuel Taropa, Hanzhao Lin, Daniel Toyama, Eliza Rutherford, Motoki Sano, HyunJeong Choe, Alex Tomala, Chalence Safranek-Shrader, Nora Kassner, Mantas Pajarskas, Matt Harvey, Sean Sechrist, Meire Fortunato, Christina Lyu, Gamaleldin Elsayed, Chenkai Kuang, James Lottes, Eric Chu, Chao Jia, Chih-Wei Chen, Peter Humphreys, Kate Baumli, Connie Tao, Rajkumar
  Samuel, Cicero~Nogueira dos Santos, Anders Andreassen, Nemanja Rakićević, Dominik Grewe, Aviral Kumar, Stephanie Winkler, Jonathan Caton, Andrew Brock, Sid Dalmia, Hannah Sheahan, Iain Barr, Yingjie Miao, Paul Natsev, Jacob Devlin, Feryal Behbahani, Flavien Prost, Yanhua Sun, Artiom Myaskovsky, Thanumalayan~Sankaranarayana Pillai, Dan Hurt, Angeliki Lazaridou, Xi~Xiong, Ce~Zheng, Fabio Pardo, Xiaowei Li, Dan Horgan, Joe Stanton, Moran Ambar, Fei Xia, Alejandro Lince, Mingqiu Wang, Basil Mustafa, Albert Webson, Hyo Lee, Rohan Anil, Martin Wicke, Timothy Dozat, Abhishek Sinha, Enrique Piqueras, Elahe Dabir, Shyam Upadhyay, Anudhyan Boral, Lisa~Anne Hendricks, Corey Fry, Josip Djolonga, Yi~Su, Jake Walker, Jane Labanowski, Ronny Huang, Vedant Misra, Jeremy Chen, RJ~Skerry-Ryan, Avi Singh, Shruti Rijhwani, Dian Yu, Alex Castro-Ros, Beer Changpinyo, Romina Datta, Sumit Bagri, Arnar~Mar Hrafnkelsson, Marcello Maggioni, Daniel Zheng, Yury Sulsky, Shaobo Hou, Tom~Le Paine, Antoine Yang, Jason Riesa, Dominika
  Rogozinska, Dror Marcus, Dalia~El Badawy, Qiao Zhang, Luyu Wang, Helen Miller, Jeremy Greer, Lars~Lowe Sjos, Azade Nova, Heiga Zen, Rahma Chaabouni, Mihaela Rosca, Jiepu Jiang, Charlie Chen, Ruibo Liu, Tara Sainath, Maxim Krikun, Alex Polozov, Jean-Baptiste Lespiau, Josh Newlan, Zeyncep Cankara, Soo Kwak, Yunhan Xu, Phil Chen, Andy Coenen, Clemens Meyer, Katerina Tsihlas, Ada Ma, Juraj Gottweis, Jinwei Xing, Chenjie Gu, Jin Miao, Christian Frank, Zeynep Cankara, Sanjay Ganapathy, Ishita Dasgupta, Steph Hughes-Fitt, Heng Chen, David Reid, Keran Rong, Hongmin Fan, Joost van Amersfoort, Vincent Zhuang, Aaron Cohen, Shixiang~Shane Gu, Anhad Mohananey, Anastasija Ilic, Taylor Tobin, John Wieting, Anna Bortsova, Phoebe Thacker, Emma Wang, Emily Caveness, Justin Chiu, Eren Sezener, Alex Kaskasoli, Steven Baker, Katie Millican, Mohamed Elhawaty, Kostas Aisopos, Carl Lebsack, Nathan Byrd, Hanjun Dai, Wenhao Jia, Matthew Wiethoff, Elnaz Davoodi, Albert Weston, Lakshman Yagati, Arun Ahuja, Isabel Gao, Golan Pundak,
  Susan Zhang, Michael Azzam, Khe~Chai Sim, Sergi Caelles, James Keeling, Abhanshu Sharma, Andy Swing, YaGuang Li, Chenxi Liu, Carrie~Grimes Bostock, Yamini Bansal, Zachary Nado, Ankesh Anand, Josh Lipschultz, Abhijit Karmarkar, Lev Proleev, Abe Ittycheriah, Soheil~Hassas Yeganeh, George Polovets, Aleksandra Faust, Jiao Sun, Alban Rrustemi, Pen Li, Rakesh Shivanna, Jeremiah Liu, Chris Welty, Federico Lebron, Anirudh Baddepudi, Sebastian Krause, Emilio Parisotto, Radu Soricut, Zheng Xu, Dawn Bloxwich, Melvin Johnson, Behnam Neyshabur, Justin Mao-Jones, Renshen Wang, Vinay Ramasesh, Zaheer Abbas, Arthur Guez, Constant Segal, Duc~Dung Nguyen, James Svensson, Le~Hou, Sarah York, Kieran Milan, Sophie Bridgers, Wiktor Gworek, Marco Tagliasacchi, James Lee-Thorp, Michael Chang, Alexey Guseynov, Ale~Jakse Hartman, Michael Kwong, Ruizhe Zhao, Sheleem Kashem, Elizabeth Cole, Antoine Miech, Richard Tanburn, Mary Phuong, Filip Pavetic, Sebastien Cevey, Ramona Comanescu, Richard Ives, Sherry Yang, Cosmo Du, Bo~Li, Zizhao
  Zhang, Mariko Iinuma, Clara~Huiyi Hu, Aurko Roy, Shaan Bijwadia, Zhenkai Zhu, Danilo Martins, Rachel Saputro, Anita Gergely, Steven Zheng, Dawei Jia, Ioannis Antonoglou, Adam Sadovsky, Shane Gu, Yingying Bi, Alek Andreev, Sina Samangooei, Mina Khan, Tomas Kocisky, Angelos Filos, Chintu Kumar, Colton Bishop, Adams Yu, Sarah Hodkinson, Sid Mittal, Premal Shah, Alexandre Moufarek, Yong Cheng, Adam Bloniarz, Jaehoon Lee, Pedram Pejman, Paul Michel, Stephen Spencer, Vladimir Feinberg, Xuehan Xiong, Nikolay Savinov, Charlotte Smith, Siamak Shakeri, Dustin Tran, Mary Chesus, Bernd Bohnet, George Tucker, Tamara von Glehn, Carrie Muir, Yiran Mao, Hideto Kazawa, Ambrose Slone, Kedar Soparkar, Disha Shrivastava, James Cobon-Kerr, Michael Sharman, Jay Pavagadhi, Carlos Araya, Karolis Misiunas, Nimesh Ghelani, Michael Laskin, David Barker, Qiujia Li, Anton Briukhov, Neil Houlsby, Mia Glaese, Balaji Lakshminarayanan, Nathan Schucher, Yunhao Tang, Eli Collins, Hyeontaek Lim, Fangxiaoyu Feng, Adria Recasens, Guangda Lai,
  Alberto Magni, Nicola~De Cao, Aditya Siddhant, Zoe Ashwood, Jordi Orbay, Mostafa Dehghani, Jenny Brennan, Yifan He, Kelvin Xu, Yang Gao, Carl Saroufim, James Molloy, Xinyi Wu, Seb Arnold, Solomon Chang, Julian Schrittwieser, Elena Buchatskaya, Soroush Radpour, Martin Polacek, Skye Giordano, Ankur Bapna, Simon Tokumine, Vincent Hellendoorn, Thibault Sottiaux, Sarah Cogan, Aliaksei Severyn, Mohammad Saleh, Shantanu Thakoor, Laurent Shefey, Siyuan Qiao, Meenu Gaba, Shuo yiin Chang, Craig Swanson, Biao Zhang, Benjamin Lee, Paul~Kishan Rubenstein, Gan Song, Tom Kwiatkowski, Anna Koop, Ajay Kannan, David Kao, Parker Schuh, Axel Stjerngren, Golnaz Ghiasi, Gena Gibson, Luke Vilnis, Ye~Yuan, Felipe~Tiengo Ferreira, Aishwarya Kamath, Ted Klimenko, Ken Franko, Kefan Xiao, Indro Bhattacharya, Miteyan Patel, Rui Wang, Alex Morris, Robin Strudel, Vivek Sharma, Peter Choy, Sayed~Hadi Hashemi, Jessica Landon, Mara Finkelstein, Priya Jhakra, Justin Frye, Megan Barnes, Matthew Mauger, Dennis Daun, Khuslen Baatarsukh, Matthew
  Tung, Wael Farhan, Henryk Michalewski, Fabio Viola, Felix de~Chaumont~Quitry, Charline~Le Lan, Tom Hudson, Qingze Wang, Felix Fischer, Ivy Zheng, Elspeth White, Anca Dragan, Jean baptiste Alayrac, Eric Ni, Alexander Pritzel, Adam Iwanicki, Michael Isard, Anna Bulanova, Lukas Zilka, Ethan Dyer, Devendra Sachan, Srivatsan Srinivasan, Hannah Muckenhirn, Honglong Cai, Amol Mandhane, Mukarram Tariq, Jack~W. Rae, Gary Wang, Kareem Ayoub, Nicholas FitzGerald, Yao Zhao, Woohyun Han, Chris Alberti, Dan Garrette, Kashyap Krishnakumar, Mai Gimenez, Anselm Levskaya, Daniel Sohn, Josip Matak, Inaki Iturrate, Michael~B. Chang, Jackie Xiang, Yuan Cao, Nishant Ranka, Geoff Brown, Adrian Hutter, Vahab Mirrokni, Nanxin Chen, Kaisheng Yao, Zoltan Egyed, Francois Galilee, Tyler Liechty, Praveen Kallakuri, Evan Palmer, Sanjay Ghemawat, Jasmine Liu, David Tao, Chloe Thornton, Tim Green, Mimi Jasarevic, Sharon Lin, Victor Cotruta, Yi-Xuan Tan, Noah Fiedel, Hongkun Yu, Ed~Chi, Alexander Neitz, Jens Heitkaemper, Anu Sinha, Denny
  Zhou, Yi~Sun, Charbel Kaed, Brice Hulse, Swaroop Mishra, Maria Georgaki, Sneha Kudugunta, Clement Farabet, Izhak Shafran, Daniel Vlasic, Anton Tsitsulin, Rajagopal Ananthanarayanan, Alen Carin, Guolong Su, Pei Sun, Shashank V, Gabriel Carvajal, Josef Broder, Iulia Comsa, Alena Repina, William Wong, Warren~Weilun Chen, Peter Hawkins, Egor Filonov, Lucia Loher, Christoph Hirnschall, Weiyi Wang, Jingchen Ye, Andrea Burns, Hardie Cate, Diana~Gage Wright, Federico Piccinini, Lei Zhang, Chu-Cheng Lin, Ionel Gog, Yana Kulizhskaya, Ashwin Sreevatsa, Shuang Song, Luis~C. Cobo, Anand Iyer, Chetan Tekur, Guillermo Garrido, Zhuyun Xiao, Rupert Kemp, Huaixiu~Steven Zheng, Hui Li, Ananth Agarwal, Christel Ngani, Kati Goshvadi, Rebeca Santamaria-Fernandez, Wojciech Fica, Xinyun Chen, Chris Gorgolewski, Sean Sun, Roopal Garg, Xinyu Ye, S.~M.~Ali Eslami, Nan Hua, Jon Simon, Pratik Joshi, Yelin Kim, Ian Tenney, Sahitya Potluri, Lam~Nguyen Thiet, Quan Yuan, Florian Luisier, Alexandra Chronopoulou, Salvatore Scellato, Praveen
  Srinivasan, Minmin Chen, Vinod Koverkathu, Valentin Dalibard, Yaming Xu, Brennan Saeta, Keith Anderson, Thibault Sellam, Nick Fernando, Fantine Huot, Junehyuk Jung, Mani Varadarajan, Michael Quinn, Amit Raul, Maigo Le, Ruslan Habalov, Jon Clark, Komal Jalan, Kalesha Bullard, Achintya Singhal, Thang Luong, Boyu Wang, Sujeevan Rajayogam, Julian Eisenschlos, Johnson Jia, Daniel Finchelstein, Alex Yakubovich, Daniel Balle, Michael Fink, Sameer Agarwal, Jing Li, Dj~Dvijotham, Shalini Pal, Kai Kang, Jaclyn Konzelmann, Jennifer Beattie, Olivier Dousse, Diane Wu, Remi Crocker, Chen Elkind, Siddhartha~Reddy Jonnalagadda, Jong Lee, Dan Holtmann-Rice, Krystal Kallarackal, Rosanne Liu, Denis Vnukov, Neera Vats, Luca Invernizzi, Mohsen Jafari, Huanjie Zhou, Lilly Taylor, Jennifer Prendki, Marcus Wu, Tom Eccles, Tianqi Liu, Kavya Kopparapu, Francoise Beaufays, Christof Angermueller, Andreea Marzoca, Shourya Sarcar, Hilal Dib, Jeff Stanway, Frank Perbet, Nejc Trdin, Rachel Sterneck, Andrey Khorlin, Dinghua Li, Xihui Wu,
  Sonam Goenka, David Madras, Sasha Goldshtein, Willi Gierke, Tong Zhou, Yaxin Liu, Yannie Liang, Anais White, Yunjie Li, Shreya Singh, Sanaz Bahargam, Mark Epstein, Sujoy Basu, Li~Lao, Adnan Ozturel, Carl Crous, Alex Zhai, Han Lu, Zora Tung, Neeraj Gaur, Alanna Walton, Lucas Dixon, Ming Zhang, Amir Globerson, Grant Uy, Andrew Bolt, Olivia Wiles, Milad Nasr, Ilia Shumailov, Marco Selvi, Francesco Piccinno, Ricardo Aguilar, Sara McCarthy, Misha Khalman, Mrinal Shukla, Vlado Galic, John Carpenter, Kevin Villela, Haibin Zhang, Harry Richardson, James Martens, Matko Bosnjak, Shreyas~Rammohan Belle, Jeff Seibert, Mahmoud Alnahlawi, Brian McWilliams, Sankalp Singh, Annie Louis, Wen Ding, Dan Popovici, Lenin Simicich, Laura Knight, Pulkit Mehta, Nishesh Gupta, Chongyang Shi, Saaber Fatehi, Jovana Mitrovic, Alex Grills, Joseph Pagadora, Tsendsuren Munkhdalai, Dessie Petrova, Danielle Eisenbud, Zhishuai Zhang, Damion Yates, Bhavishya Mittal, Nilesh Tripuraneni, Yannis Assael, Thomas Brovelli, Prateek Jain, Mihajlo
  Velimirovic, Canfer Akbulut, Jiaqi Mu, Wolfgang Macherey, Ravin Kumar, Jun Xu, Haroon Qureshi, Gheorghe Comanici, Jeremy Wiesner, Zhitao Gong, Anton Ruddock, Matthias Bauer, Nick Felt, Anirudh GP, Anurag Arnab, Dustin Zelle, Jonas Rothfuss, Bill Rosgen, Ashish Shenoy, Bryan Seybold, Xinjian Li, Jayaram Mudigonda, Goker Erdogan, Jiawei Xia, Jiri Simsa, Andrea Michi, Yi~Yao, Christopher Yew, Steven Kan, Isaac Caswell, Carey Radebaugh, Andre Elisseeff, Pedro Valenzuela, Kay McKinney, Kim Paterson, Albert Cui, Eri Latorre-Chimoto, Solomon Kim, William Zeng, Ken Durden, Priya Ponnapalli, Tiberiu Sosea, Christopher~A. Choquette-Choo, James Manyika, Brona Robenek, Harsha Vashisht, Sebastien Pereira, Hoi Lam, Marko Velic, Denese Owusu-Afriyie, Katherine Lee, Tolga Bolukbasi, Alicia Parrish, Shawn Lu, Jane Park, Balaji Venkatraman, Alice Talbert, Lambert Rosique, Yuchung Cheng, Andrei Sozanschi, Adam Paszke, Praveen Kumar, Jessica Austin, Lu~Li, Khalid Salama, Bartek Perz, Wooyeol Kim, Nandita Dukkipati, Anthony
  Baryshnikov, Christos Kaplanis, XiangHai Sheng, Yuri Chervonyi, Caglar Unlu, Diego de~Las~Casas, Harry Askham, Kathryn Tunyasuvunakool, Felix Gimeno, Siim Poder, Chester Kwak, Matt Miecnikowski, Vahab Mirrokni, Alek Dimitriev, Aaron Parisi, Dangyi Liu, Tomy Tsai, Toby Shevlane, Christina Kouridi, Drew Garmon, Adrian Goedeckemeyer, Adam~R. Brown, Anitha Vijayakumar, Ali Elqursh, Sadegh Jazayeri, Jin Huang, Sara~Mc Carthy, Jay Hoover, Lucy Kim, Sandeep Kumar, Wei Chen, Courtney Biles, Garrett Bingham, Evan Rosen, Lisa Wang, Qijun Tan, David Engel, Francesco Pongetti, Dario de~Cesare, Dongseong Hwang, Lily Yu, Jennifer Pullman, Srini Narayanan, Kyle Levin, Siddharth Gopal, Megan Li, Asaf Aharoni, Trieu Trinh, Jessica Lo, Norman Casagrande, Roopali Vij, Loic Matthey, Bramandia Ramadhana, Austin Matthews, CJ~Carey, Matthew Johnson, Kremena Goranova, Rohin Shah, Shereen Ashraf, Kingshuk Dasgupta, Rasmus Larsen, Yicheng Wang, Manish~Reddy Vuyyuru, Chong Jiang, Joana Ijazi, Kazuki Osawa, Celine Smith, Ramya~Sree
  Boppana, Taylan Bilal, Yuma Koizumi, Ying Xu, Yasemin Altun, Nir Shabat, Ben Bariach, Alex Korchemniy, Kiam Choo, Olaf Ronneberger, Chimezie Iwuanyanwu, Shubin Zhao, David Soergel, Cho-Jui Hsieh, Irene Cai, Shariq Iqbal, Martin Sundermeyer, Zhe Chen, Elie Bursztein, Chaitanya Malaviya, Fadi Biadsy, Prakash Shroff, Inderjit Dhillon, Tejasi Latkar, Chris Dyer, Hannah Forbes, Massimo Nicosia, Vitaly Nikolaev, Somer Greene, Marin Georgiev, Pidong Wang, Nina Martin, Hanie Sedghi, John Zhang, Praseem Banzal, Doug Fritz, Vikram Rao, Xuezhi Wang, Jiageng Zhang, Viorica Patraucean, Dayou Du, Igor Mordatch, Ivan Jurin, Lewis Liu, Ayush Dubey, Abhi Mohan, Janek Nowakowski, Vlad-Doru Ion, Nan Wei, Reiko Tojo, Maria~Abi Raad, Drew~A. Hudson, Vaishakh Keshava, Shubham Agrawal, Kevin Ramirez, Zhichun Wu, Hoang Nguyen, Ji~Liu, Madhavi Sewak, Bryce Petrini, DongHyun Choi, Ivan Philips, Ziyue Wang, Ioana Bica, Ankush Garg, Jarek Wilkiewicz, Priyanka Agrawal, Xiaowei Li, Danhao Guo, Emily Xue, Naseer Shaik, Andrew Leach,
  Sadh~MNM Khan, Julia Wiesinger, Sammy Jerome, Abhishek Chakladar, Alek~Wenjiao Wang, Tina Ornduff, Folake Abu, Alireza Ghaffarkhah, Marcus Wainwright, Mario Cortes, Frederick Liu, Joshua Maynez, Andreas Terzis, Pouya Samangouei, Riham Mansour, Tomasz Kępa, François-Xavier Aubet, Anton Algymr, Dan Banica, Agoston Weisz, Andras Orban, Alexandre Senges, Ewa Andrejczuk, Mark Geller, Niccolo~Dal Santo, Valentin Anklin, Majd~Al Merey, Martin Baeuml, Trevor Strohman, Junwen Bai, Slav Petrov, Yonghui Wu, Demis Hassabis, Koray Kavukcuoglu, Jeff Dean, and Oriol Vinyals.
\newblock Gemini 1.5: Unlocking multimodal understanding across millions of tokens of context, 2024.
\newblock URL \url{https://arxiv.org/abs/2403.05530}.

\bibitem[Tian et~al.(2023)Tian, Mitchell, Yao, Manning, and Finn]{tian2023finetuning}
Katherine Tian, Eric Mitchell, Huaxiu Yao, Christopher~D. Manning, and Chelsea Finn.
\newblock Fine-tuning language models for factuality, 2023.

\bibitem[Van~Veen et~al.(2024)Van~Veen, Van~Uden, Blankemeier, Delbrouck, Aali, Bluethgen, Pareek, Polacin, Reis, Seehofnerová, Rohatgi, Hosamani, Collins, Ahuja, Langlotz, Hom, Gatidis, Pauly, and Chaudhari]{Van_Veen_2024}
Dave Van~Veen, Cara Van~Uden, Louis Blankemeier, Jean-Benoit Delbrouck, Asad Aali, Christian Bluethgen, Anuj Pareek, Malgorzata Polacin, Eduardo~Pontes Reis, Anna Seehofnerová, Nidhi Rohatgi, Poonam Hosamani, William Collins, Neera Ahuja, Curtis~P. Langlotz, Jason Hom, Sergios Gatidis, John Pauly, and Akshay~S. Chaudhari.
\newblock Adapted large language models can outperform medical experts in clinical text summarization.
\newblock \emph{Nature Medicine}, 30\penalty0 (4):\penalty0 1134–1142, February 2024.
\newblock ISSN 1546-170X.
\newblock \doi{10.1038/s41591-024-02855-5}.
\newblock URL \url{http://dx.doi.org/10.1038/s41591-024-02855-5}.

\bibitem[Wadden et~al.(2020)Wadden, Lo, Wang, Lin, van Zuylen, Cohan, and Hajishirzi]{Wadden2020FactOF}
David Wadden, Kyle Lo, Lucy~Lu Wang, Shanchuan Lin, Madeleine van Zuylen, Arman Cohan, and Hannaneh Hajishirzi.
\newblock Fact or fiction: Verifying scientific claims.
\newblock In \emph{Conference on Empirical Methods in Natural Language Processing}, 2020.
\newblock URL \url{https://api.semanticscholar.org/CorpusID:216867133}.

\bibitem[Wadhwa et~al.(2023)Wadhwa, DeYoung, Nye, Amir, and Wallace]{wadhwa2023jointly}
Somin Wadhwa, Jay DeYoung, Benjamin Nye, Silvio Amir, and Byron~C Wallace.
\newblock Jointly extracting interventions, outcomes, and findings from rct reports with llms.
\newblock In \emph{Machine Learning for Healthcare Conference}, pages 754--771. PMLR, 2023.

\bibitem[Wanner et~al.(2024)Wanner, Ebner, Jiang, Dredze, and Van~Durme]{wanner-etal-2024-closer}
Miriam Wanner, Seth Ebner, Zhengping Jiang, Mark Dredze, and Benjamin Van~Durme.
\newblock A closer look at claim decomposition.
\newblock In Danushka Bollegala and Vered Shwartz, editors, \emph{Proceedings of the 13th Joint Conference on Lexical and Computational Semantics (*SEM 2024)}, pages 153--175, Mexico City, Mexico, June 2024. Association for Computational Linguistics.
\newblock \doi{10.18653/v1/2024.starsem-1.13}.
\newblock URL \url{https://aclanthology.org/2024.starsem-1.13}.

\bibitem[Williams et~al.(2018)Williams, Nangia, and Bowman]{williams-etal-2018-broad}
Adina Williams, Nikita Nangia, and Samuel Bowman.
\newblock A broad-coverage challenge corpus for sentence understanding through inference.
\newblock In Marilyn Walker, Heng Ji, and Amanda Stent, editors, \emph{Proceedings of the 2018 Conference of the North {A}merican Chapter of the Association for Computational Linguistics: Human Language Technologies, Volume 1 (Long Papers)}, pages 1112--1122, New Orleans, Louisiana, June 2018. Association for Computational Linguistics.
\newblock \doi{10.18653/v1/N18-1101}.
\newblock URL \url{https://aclanthology.org/N18-1101}.

\bibitem[Wornow et~al.(2024)Wornow, Lozano, Dash, Jindal, Mahaffey, and Shah]{wornow2024zeroshot}
Michael Wornow, Alejandro Lozano, Dev Dash, Jenelle Jindal, Kenneth~W. Mahaffey, and Nigam~H. Shah.
\newblock Zero-shot clinical trial patient matching with llms, 2024.

\bibitem[Wright et~al.(2022)Wright, Wadden, Lo, Kuehl, Cohan, Augenstein, and Wang]{Wright2022GeneratingSC}
Dustin Wright, David Wadden, Kyle Lo, Bailey Kuehl, Arman Cohan, Isabelle Augenstein, and Lucy~Lu Wang.
\newblock Generating scientific claims for zero-shot scientific fact checking.
\newblock \emph{ArXiv}, abs/2203.12990, 2022.
\newblock URL \url{https://api.semanticscholar.org/CorpusID:247627737}.

\bibitem[Xie et~al.(2024)Xie, Zhang, Cheng, Liu, Gero, Wong, Naumann, Poon, and Rose]{xie-etal-2024-doclens}
Yiqing Xie, Sheng Zhang, Hao Cheng, Pengfei Liu, Zelalem Gero, Cliff Wong, Tristan Naumann, Hoifung Poon, and Carolyn Rose.
\newblock {D}oc{L}ens: Multi-aspect fine-grained medical text evaluation.
\newblock In Lun-Wei Ku, Andre Martins, and Vivek Srikumar, editors, \emph{Proceedings of the 62nd Annual Meeting of the Association for Computational Linguistics (Volume 1: Long Papers)}, pages 649--679, Bangkok, Thailand, August 2024. Association for Computational Linguistics.
\newblock \doi{10.18653/v1/2024.acl-long.39}.
\newblock URL \url{https://aclanthology.org/2024.acl-long.39}.

\bibitem[Yang et~al.(2024)Yang, Liu, Deng, Wu, Weng, Zhou, and Wang]{yang2024enhancing}
Jingye Yang, Cong Liu, Wendy Deng, Da~Wu, Chunhua Weng, Yunyun Zhou, and Kai Wang.
\newblock Enhancing phenotype recognition in clinical notes using large language models: Phenobcbert and phenogpt.
\newblock \emph{Patterns}, 5\penalty0 (1), 2024.

\bibitem[Zhang et~al.(2020)Zhang, Kishore, Wu, Weinberger, and Artzi]{zhang2020bertscoreevaluatingtextgeneration}
Tianyi Zhang, Varsha Kishore, Felix Wu, Kilian~Q. Weinberger, and Yoav Artzi.
\newblock Bertscore: Evaluating text generation with bert, 2020.
\newblock URL \url{https://arxiv.org/abs/1904.09675}.

\bibitem[Zheng et~al.(2023)Zheng, Chiang, Sheng, Zhuang, Wu, Zhuang, Lin, Li, Li, Xing, et~al.]{zheng2023judging}
Lianmin Zheng, Wei-Lin Chiang, Ying Sheng, Siyuan Zhuang, Zhanghao Wu, Yonghao Zhuang, Zi~Lin, Zhuohan Li, Dacheng Li, Eric Xing, et~al.
\newblock Judging llm-as-a-judge with mt-bench and chatbot arena.
\newblock \emph{Advances in Neural Information Processing Systems}, 36:\penalty0 46595--46623, 2023.

\end{thebibliography}

\newpage
\appendix

\section{Dataset Details}
  
\subsection{Data Sources for Clinical Notes}
\label{apx:dataset_sources}

For \datasetname, we sampled 2,168 clinical notes from three de-identified research datasets. We describe each of these datasets in detail and summarize the number of notes from each of them in table~\ref{tab:data_statistics} and token statistics in table~\ref{tab:tokens}. We describe each source in detail below:

\paragraph{MIMIC} MIMIC-III \citep{Johnson2016MIMICIIIAF} is a database of de-identified EHR comprising over 40k patients admitted to the intensive care unit of the Beth Israel Deaconess Medical Center between 2001 and 2012.
MIMIC-III includes several note types, we sample \textbf{progress notes, discharge summaries and  nursing notes}.


\paragraph{MIMIC-CXR} The MIMIC Chest X-ray Database v2.0.0 \citep{johnson2019mimic} is a large publicly available dataset of chest radiographs in DICOM format with free-text radiology reports. The dataset contains 377,110 images corresponding to 227,835 radiographic studies performed at the Beth Israel Deaconess Medical Center in Boston, MA. We only use the free-text \textbf{radiology reports} for fact generation. We mapped the test splits from MIMIC-CXR with MIMIC-III and used a sub-sample of the split for our experiments.

\paragraph{CORAL} is a collection of a diverse set of 20 breast cancer and 20 pancreatic cancer patients from the University of California, San Francisco (UCSF) Information Commons, which contained patient data between 2012–2022, de-identified with Philter \citep{article}.
We use all the 172 progress notes for our experiments.

\paragraph{MedAlign} is collection of de-identified EHR data from Stanford Hospital and Lucile Packard Children’s Hospital. We sub-sample \textbf{progress notes, nursing notes and discharge summaries} from MedAlign \citep{fleming2023medalign} which consists of 276 longitudinal EHRs, out of which we use 750 notes.

\begin{table}[ht!]
\centering
\small
\begin{tabular}{l@{\hskip 6pt}c@{\hskip 6pt}c@{\hskip 6pt}c}
\toprule
\multirow{2}{*}{\textbf{Note Type}} & \multicolumn{3}{c}{\textbf{\# of Notes}} \\ \cmidrule{2-4}
                           & MIMIC & MedAlign & CORAL \\ \midrule
Progress Note              & 250   & 250 & 172\\
Nursing Note               & 250   & 129 & -    \\
Discharge Summary          & 250   & 117 & -    \\ 
Procedure Note             & -     & 250 & -    \\ 
-- \textit{Radiology Note }& 500   & - & -    \\ \midrule
\textbf{Total}             & \textbf{1250} & \textbf{750} & \textbf{172} \\ \bottomrule
\end{tabular}
\caption{Counts of notes by type and data source. Radiology notes are a subtype of procedure notes, and were sampled from both MIMIC-III and MIMIC-CXR.}
\label{tab:data_statistics}
\end{table}

\begin{table}[ht!]
\centering
\small
\begin{tabular}{lcccc}
\toprule
\textbf{Note Type} & \multicolumn{2}{c}{\textbf{\# Tokens}} \\
\cmidrule{2-3}
 & Mean (SD) & Min/Max \\
\midrule
Progress Note      & 1510 (1199) & [ 65, 6758 ]  \\
Nursing Note       & 247 (168)   & [ 60, 977 ]   \\
Discharge Summary  & 2253 (1041) & [ 155, 7264 ] \\
Procedure Notes        & 339 (314)   & [ 62, 3047 ]  \\
\bottomrule
\end{tabular}
\caption{Note token length summary statistics.}
\label{tab:tokens}

\end{table}

\subsection{Entailment Data Sources}
\label{apx:NLI_datasets}

For entailment evaluation, we use existing NLI datasets. We describe them in detail, in the following paragraphs:

\paragraph{MultiNLI} \citep{williams-etal-2018-broad} The Multi-Genre Natural Language Inference corpus is a crowd-sourced collection of 433k sentence pairs annotated with textual entailment information. The labels include \textit{entailment}, \textit{neutral} and \textit{contradiction}.

\paragraph{MedNLI} \citep{romanov2018lessonsnaturallanguageinference} is a dataset annotated by healthcare professionals for the task of natural language inference (NLI) based on patient medical histories. The premise sentences in this dataset are sourced from the MIMIC-III \citep{Johnson2016MIMICIIIAF} database. They use \textit{Past Medical History section} where annotators write alternative sentence which could be one of these  \textit{entailment}, \textit{neutral} or a \textit{contradiction}. The \textit{test} split consists of 1422 such pairs along with the label.

\paragraph{Scitail} \citep{DBLP:conf/aaai/KhotSC18} The SCITAIL dataset is created for answering school-level science questions by converting each question and its correct answer into an assertive hypothesis (H). Relevant sentences are extracted from a large text corpus to serve as premises (P). Each premise-hypothesis pair is then annotated (via crowdsourcing) as entails or neutral. There are 2,126 pairs in the test split.

\paragraph{SNLI} \citep{bowman-etal-2015-large} The Stanford Natural Language Inference corpusis a collection of 570k human-written English sentence pairs manually labeled for balanced classification with the labels entailment, contradiction, and neutral. Their test split is 10,000 sentence pairs.

\section{Note Types}
\label{apx:note_types}

\datasetname~ includes samples from four clinical note types.
We outline the types and their clinical purpose below.

\paragraph{Procedure Note} Document medical procedures, including diagnostic and therapeutic interventions. Radiology reports, as a key example, detail findings from imaging studies like X-rays, CT scans, and MRIs, often comparing them to prior exams. Other examples include endoscopies, biopsies, and surgical procedures. These notes typically include the procedure performed, clinical indications, findings, and follow-up recommendations.

\paragraph{Nursing Note} Written by nurses to provide a systematic assessment of a patient’s condition across individual body systems (e.g., cardiovascular, neurologic) during a specific time frame. 
The focus of the note is primarily on the patient’s current status, with less emphasis on detailed planning for future care.

\paragraph{Progress Note} Written by physicians to summarize a patient’s medical status over the preceding day and outline the care plan for the following day. 
They include a review of significant events and relevant diagnostic tests conducted during that period. They include key events, diagnostic tests, a current patient exam, active medical issues.

\paragraph{Discharge Summary} Written by physicians to synthesize key medical information from a patient's hospitalization, including clinical notes, diagnostic reports, and treatment plans. They provide a concise overview of the patient’s presentation, past medical history, key findings, future medical plans, and discharge medications for subsequent healthcare providers.

\section{Choice of Models}
\label{apx:models}

While we acknowledge the potential value of domain-specific models, numerous studies have shown that general-domain models often outperform medically pretrained\citep{jeong2024medicaladaptationlargelanguage, Van_Veen_2024, ceballos-arroyo-etal-2024-open}, or fine-tuned models across a range of tasks such classification, information extraction, and summarization, including our initial experiments.

\begin{table}[ht]
\centering
\small
\begin{tabular}{llc} 
\toprule \textbf{Note Type} & \textbf{MedLM }   & \textbf{Gemini-1.5 }           \\ 
\midrule
Discharge summary  &  58.5 & 64 \\ 
Nursing note & 77.4 & 79.8 \\ 
Progress note  &   74.8    & 77.5   \\
\bottomrule
\end{tabular}
\caption{Comparison of scores of MedLM and Gemini. We report Fact-F1 scores. \textbf{Gemini-1.5 consistently outperformed MedLM} in our evaluations.}
\label{tab:medlm-gemini}
\end{table}

\section{Hyperparameters}
\label{sec:hyperparameters}
For the fact decomposition task, we evaluated four models. GPT-4o, Gemini Flash 1.5, and LLaMA 3 8B were configured with a maximum token generation length of 4000 tokens, temperature of 0.01 and a top-p value of 0.9. Additionally, we included the o1-mini model, which used a temperature of 1, a top-p value of 1, and a maximum of 16,000 new tokens to allow for extra tokens generated in its chain of thought reasoning. At the time of our experiments o1-mini did not support other temperature or top-p settings. For the entailment task, we evaluated five models: GPT-4o, Gemini Flash 1.5, LLaMA 3 8B, LLaMA 3 70B, and GPT-4o-mini. These models were configured with a maximum token generation length of 25 tokens, with the same generation parameters as the fact decomposition task: a temperature of 0.01 and a top-p value of 0.9.

\section{Entailment Annotation Guidelines}
\label{apx:annotation_guidelines}

The following instructions were provided to clinical annotators for evaluating entailment of premise-hypothesis pairs.

\subsection*{Annotation Instructions}
\begin{enumerate}
    \item You have two sheets: one containing the \textbf{reference texts} (labeled as 1 and 2) and one for \textbf{annotating claims}.
    \item Start in the \textbf{annotation} sheet.
    \item For each row, review the text in the \texttt{claim} column and determine if all the information in it can be \textbf{completely} inferred from the reference text specified in the \texttt{reference\_ID} column.
    \item Annotate as follows:
    \begin{itemize}
        \item \textbf{1}: If all of the information in the \texttt{claim} column can be \textbf{completely} inferred from the reference text.
        \item \textbf{0}: If any part of the \texttt{claim} contains information that is not present in the reference text.
    \end{itemize}
    \item If necessary, open the reference text sheet and annotation sheet in two separate windows for ease of comparison.
\end{enumerate}

\subsection*{Examples}
\textbf{Example of a claim that can be completely inferred (annotation = 1):}  
\textbf{Reference text:} Left pleural effusion increased from prior scan.  
\textbf{Claim:} There is a pleural effusion on the left side.  
\textbf{Explanation:} The claim is completely inferable from the reference text.

\textbf{Example of a claim that can be completely inferred (annotation = 1):}  
\textbf{Reference text:} FINDINGS: Nearly complete opacification of the left hemithorax is of increasing density since the recent prior CT.  
\textbf{Claim:} The results showed nearly complete opacification of the left hemithorax, which has increased in density since the last CT scan.  
\textbf{Explanation:} The claim is completely inferable from the reference text.

\textbf{Example of a claim that cannot be completely inferred (annotation = 0):}  
\textbf{Reference text:} Left pleural effusion increased from prior scan.  
\textbf{Claim:} The CT scan shows that the pleural effusion is increased from the prior scan.  
\textbf{Explanation:} There is no mention of a CT scan in the reference text.

\textbf{Example of a claim that cannot be completely inferred (annotation = 0):}  
\textbf{Reference text:} The patient has been evaluated for potential consolidation or pneumothorax.  
Multiple prior chest radiographs and a recent Chest CT were referred for comparison.  
\textbf{Claim:} Evaluate for evidence of consolidation or pneumothorax.  
COMPARISON: Multiple prior chest radiographs most recent on 1/15/22.  
\textbf{Explanation:} The reference text does not mention that the most recent scan was on 1/15/22.

\textbf{Example of a claim that cannot be completely inferred (annotation = 0):}  
\textbf{Reference text:}  
1. The patient is intubated and has hypoxic respiratory failure.  
2. A comparison is made with a prior study.  
3. The cardiac size is normal.  
4. The lines and tubes are in the standard position.  
5. There are large right and moderate left pleural effusions that are grossly unchanged.  
6. The right upper lobe opacity has improved consistent with improving atelectasis.  
7. The pleural effusions are associated with atelectasis and are larger on the right side.  
8. There is mild vascular congestion.  
Multiple prior chest radiographs and a recent Chest CT were referred for comparison.  
\textbf{Claim:} FINAL REPORT SINGLE FRONTAL VIEW OF THE CHEST REASON FOR EXAM: Intubated patient, hypoxic respiratory failure.  
\textbf{Explanation:} The claim contains several pieces of information that are not inferable from the reference text:  
1) The fact that it is a final report,  
2) It is a single frontal view,  
3) The reason for the exam is hypoxic respiratory failure.

\section{Qualitative Annotation Guidelines}
\label{apx:qual_anno_guidelines}

Expert reviewers assessed overall completeness of fact
decompositions and the correctness, independence, and atomicity of individual facts. 

\paragraph{Completeness: The set of fact decompositions are defined complete if the facts capture all information present in the source document.}

\begin{enumerate}
    \item: \textbf{1}: If all of the information in the \texttt{note} is covered in the \textbf{decomposed fact list} 
    \item: \textbf{0} if any part of the \texttt{note} is missed by decomposed facts.
    \item Example: 
    \texttt{Note: "No evidence of pneumothorax. \\
    Left pleural effusion increased from prior scan."\\   
    Facts: "No pneumothorax. There is a pleural effusion." } \\
    Missing: \texttt{"Pleural effusion is increased from last scan."} \\
    Score: {0}
\end{enumerate}

For individual facts, reviewers judge on three criterion: 

\paragraph{Atomicity: A fact is atomic if it represented a minimal unit of information which cannot be decomposed further.}

\begin{enumerate}
    \item \textbf{1}: The sentence presents exactly one fact, statement, or concept without unnecessary complexity.
    \item 0: The statement includes multiple facts, making it difficult to evaluate or extract key information.
    \item Example: \texttt{Note: "No evidence of pneumothorax. \\
    Left pleural effusion increased from prior scan." \\
    Facts: "Pleural effusion is left sided and increased from last scan." } \\
    Score: {0} 
\end{enumerate}

\paragraph{Correctness: A fact is correct if it accurately reflects the contents of the source.}

\begin{enumerate}
    \item \textbf{1}: The fact presented accurately reflects the information in the note.
    \item \textbf{0}: The statement includes multiple facts, making it difficult to evaluate or extract key information.
    \item Example: \texttt{Note: "No evidence of pneumothorax.\\
    Left pleural effusion increased from prior scan."\\
    Facts: "There is a large pneumothorax." }\\
    Score: {0}
\end{enumerate}

\paragraph{Independence: A fact is independent if it can be understood on its own.}

\begin{enumerate}
    \item \textbf{1}: The fact presented accurately reflects the information in the note.
    \item \textbf{0}: The statement includes multiple facts, making it difficult to evaluate or extract key information.
    \item Example: \texttt{Note: "No evidence of pneumothorax.\\
    Left pleural effusion increased from prior scan."\\
    Facts: "It is left sided."}\\
    Score: {0}
\end{enumerate}

\section{Compute Environment}

Experiments are performed in a HIPAA compliant APIs and local on-prem university compute environment using 4 Nvidia H100 GPUs. All compute environments supported HIPAA-compliant data protocols.


\section{Sentence Tokenization}

Sentence tokenization errors resulting from formatting characters and tabular data are show in Table \ref{tbl:senterrors}.

\begin{table*}[ht!]
\small
\centering
\begin{tabular}{@{}lrrp{8cm}@{}}
\toprule
\textbf{Note Type}       & \textbf{Junk} & \textbf{Partial} & \textbf{Examples} \\ \midrule
Discharge Summary       & 1.2\%              & 3.6\%                 & \raggedright\arraybackslash\colorbox[gray]{0.9}{\texttt{.}}, \colorbox[gray]{0.9}{\texttt{Rel.}}, \colorbox[gray]{0.9}{\texttt{05:55}}, \colorbox[gray]{0.9}{\texttt{Trimethoprim/Sulfamethoxazole.}}, \colorbox[gray]{0.9}{\texttt{SIRS/Sepsis}}, \colorbox[gray]{0.9}{\texttt{2+}}, \colorbox[gray]{0.9}{\texttt{05/21/2018}}, \colorbox[gray]{0.9}{\texttt{5}} \\
Progress Note           & 0.7\%              & 10.4\%                & \raggedright\arraybackslash\colorbox[gray]{0.9}{\texttt{In}}, \colorbox[gray]{0.9}{\texttt{Unknown;}}, \colorbox[gray]{0.9}{\texttt{jpg]}}, \colorbox[gray]{0.9}{\texttt{6.}}, \colorbox[gray]{0.9}{\texttt{101 mEq/L}}, \colorbox[gray]{0.9}{\texttt{fevers.}}, \colorbox[gray]{0.9}{\texttt{24 hours}}, \colorbox[gray]{0.9}{\texttt{28.6}}, \colorbox[gray]{0.9}{\texttt{11/23/2015}} \\
Nursing Note            & 0.2\%              & 3.7\%                 & \raggedright\arraybackslash\colorbox[gray]{0.9}{\texttt{Learning Preference}}, \colorbox[gray]{0.9}{\texttt{MAE's.}}, \colorbox[gray]{0.9}{\texttt{Pink/ruddy.}}, \colorbox[gray]{0.9}{\texttt{01/09/2016}}, \colorbox[gray]{0.9}{\texttt{PT.}}, \colorbox[gray]{0.9}{\texttt{NPN}}, \colorbox[gray]{0.9}{\texttt{AFSF.}}, \colorbox[gray]{0.9}{\texttt{bs:ronchi.}}, \colorbox[gray]{0.9}{\texttt{50}} \\
Procedures              & 9.0\%              & 6.0\%                 & \raggedright\arraybackslash\colorbox[gray]{0.9}{\texttt{\_\_\_}}, \colorbox[gray]{0.9}{\texttt{\_\_\_\_\_\_\_\_\_}}, \colorbox[gray]{0.9}{\texttt{I.}}, \colorbox[gray]{0.9}{\texttt{*}}, \colorbox[gray]{0.9}{\texttt{(Over)}}, \colorbox[gray]{0.9}{\texttt{$\mid$}}, \colorbox[gray]{0.9}{\texttt{\_\_\_\_\_\_\_}} \\ \bottomrule
\end{tabular}
\caption{Example sentence tokenization errors with separate \textit{junk} and \textit{partial} error rates. Junk sentences are formatting characters (e.g. lines or tokens denoting split sections). Partial indicates an incomplete sentence that was incorrectly split during sentence parsing.}
\label{tbl:senterrors}
\end{table*}

\section{Supplementary Results}
\label{apx:supp_res}

\subsection{Validating LLM-as-a-Judge}

A key objective of this work was to evaluate the ``LLM-as-a-judge" component within a fact decomposition verification framework. Despite the critical role of this initial step in fact verification pipelines, it has not been rigorously benchmarked in clinical text or across diverse clinical document types. The scale of human-annotated labels used in this study was driven by the need to validate our proposed NLI-based LLM-as-a-judge. To determine an appropriate sample size for these human-provided labels, we performed power calculations. These calculations ensured that our manually labeled dataset was sufficiently powered to accurately estimate sensitivity, specificity, and other performance metrics, using parameters derived from NLI benchmarking results on the FactEHR development set.

With these parameters, annotating 250 entailment pairs per note type was sufficient to provide high-confidence estimates (99\% confidence level overall and ~80\% confidence level per note type). This resulted in a total of 1000 uniquely annotated entailment pairs across the four note types. Additionally, we included 200 duplicate annotations (50 per note type) to assess inter-annotator agreement, ensuring we could reliably estimate kappa with high confidence. This brought the total for this annotation tranche to 1200 entailment pairs, including both unique and duplicate annotations. Combined with the 2468 entailment pairs already annotated, our dataset now comprises approximately 3500 human-annotated entailment pairs, representing one of the largest datasets of its kind in the clinical domain.

\begin{table}[h!]
\centering
\begin{small}
\begin{tabular}{lc}
\toprule
 \textbf{Metric} & \textbf{Performance} \\
\midrule
Expected sensitivity & 95\%   \\
Expected specificity & 87\%   \\
Prevalence of entailed claims & 70\%   \\
\bottomrule
\end{tabular}
\end{small}
\caption{Performance of  LLM-as-a-Judge model on the  250 entailment pairs per note type, they provide high-confidence estimates.}
\label{tab:validating_llm_as_a_judge}
\end{table}

\begin{table*}[ht!]
\centering
\small
\begin{tabular}{p{2cm}p{2cm}p{1cm}p{2cm}p{1.5cm}p{1.5cm}p{1.5cm}c}
\toprule
\textbf{Dataset} & \textbf{Model} &\textbf{N} & \textbf{Unparsable} &\textbf{P} & \textbf{R*} & \textbf{F1} & \textbf{Acc.} \\
\midrule
\multirow{3}{*}{FactEHR} & GPT-4o & 2468 & 0 & 90.8 & 95.7 & \textbf{93.2} & \textbf{90.0} \\
 & Gemini-1.5 & 2369 & 0 &  89.1 & 96.5 & 92.6 & 89.0 \\
 & Llama3-8 & 2468 & 0.36 &  86.2 & 44.7 & 58.8 & 55.2 \\
 & Llama3-70 & 2468 & 0.166 &  88.18 & 90.12 & 89.14 & 84.24 \\
 & 4o-mini & 2468 & 0 &  92.16 & 90.97 & 91.56 & 87.97 \\
\midrule
\multirow{3}{*}{MedNLI} & GPT-4o & 1422 & 0 & 94.7 & 76.0 & \textbf{84.3} & \textbf{90.6} \\
 & Gemini-1.5 & 1422 & 0  &  93.8 & 67.5 & 78.5 & 87.7 \\
 & Llama3-8 & 1422 & 0.001  &  65.2 & 86.7 & 74.5 & 80.2 \\
 & Llama3-70 & 1422 & 0.004 &  80.93 & 87.76 & 84.21 & 89.03 \\
 & 4o-mini & 1422 & 0 &  91.71 & 74.68 & 82.33 & 89.31 \\
  
\midrule
\multirow{3}{*}{SciTail} & GPT-4o & 2126 &  0 & 89.7 & 83.6 & \textbf{86.5} & \textbf{89.7} \\
 & Gemini-1.5 & 2120 & 0 & 79.5 & 88.2 & 83.6 & 86.3 \\
 & Llama3-8  & 2126 & 0.002 & 47.9 & 98.8 & 64.6 & 57.0 \\
 & Llama3-70  & 2126 & 0.0117 & 70.49 & 93.35 & 80.33 & 81.89 \\
  & 4o-mini & 2126 & 0 & 80 & 90.74 & 85.03 & 87.35 \\
\midrule
\multirow{3}{*}{MultiNLI} & GPT-4o & 9832  & 0 &  94.1 & 82.0 & 87.7 & 91.9 \\
 & Gemini-1.5 & 9832 & 0 &  91.4 & 86.6 & \textbf{88.9} & \textbf{92.4} \\
 & Llama3-8 & 9832 & 0.008 &  55.5 & 97.3 & 70.7 & 71.6 \\
  & Llama3-70 & 9832 & 0.007 &  79.93 & 92.23 & 85.64 & 89.11 \\
  & 4o-mini & 9832 & 0 & 90.87 & 86.49 & 88.62 & 92.18 \\
\midrule
\multirow{3}{*}{SNLI} & GPT-4o & 10000 & 0 & 93.1 & 87.0 & \textbf{90.0} & \textbf{93.4} \\
 & Gemini-1.5 & 10000 & 0 & 93.0 & 86.1 & 89.4 & 93.1 \\
 & Llama3-8 & 10000 & 0.0036 & 56.5 & 97.6 & 71.6 & 73.9 \\
 & Llama3-70 & 10000 & 0.006 & 74.08 & 94.27 & 82.96 & 86.96 \\
  & 4o-mini & 10000 & 0 & 91.77 & 86.1 & 88.85 & 92.72 \\
\bottomrule
\end{tabular}
\caption{Entailment evaluation benchmarking results. Note that Gemini-1.5 flash failed to process some examples due to content moderation (``N" column). Responses whose JSON output were not parsable were coerced to ``0". \textbf{Bolded} are the highest numbers (F1 and accuracy) of a model for each dataset. Our results show that GPT-4o performs well on four out of five benchmarking datasets, including the FactEHR validation set. Therefore, we selected GPT-4o as the final LLM entailment judge.}
\label{tab:NLI_benchmarks}
\end{table*}

\subsection{Fact Atomicity}

\begin{figure}[h]
    \centering
    {\includegraphics[width=0.48\textwidth]{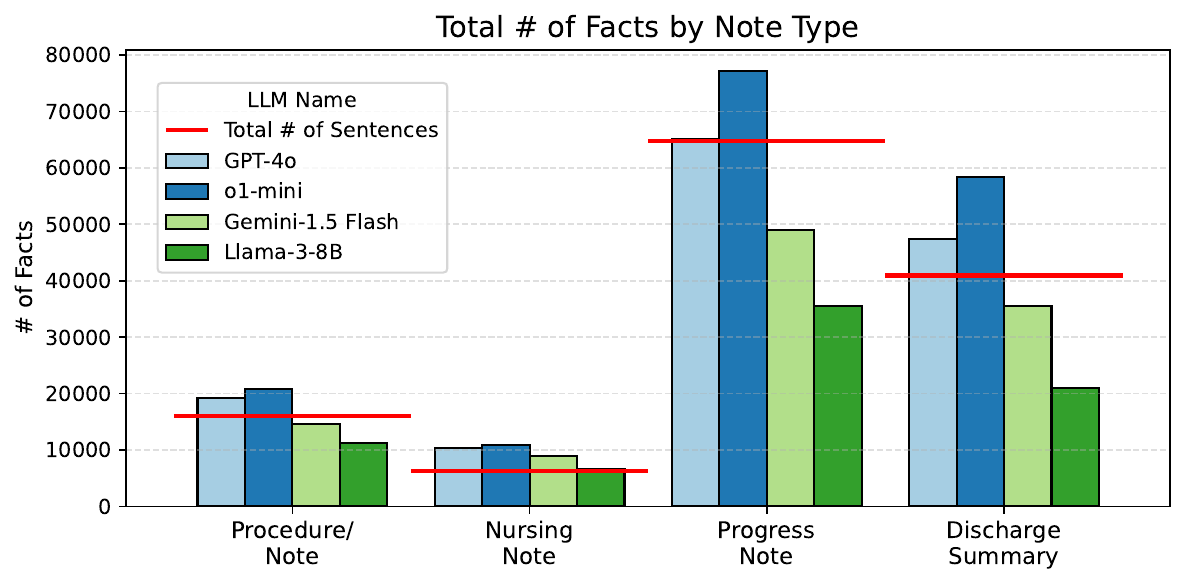}}
    \caption{The total number of generated facts varies significantly across LLMs and note types. The red line indicates the total sentence count per note type.} 
    \label{fig:total_facts}
\end{figure}

We plot the total number of generated facts in figure~\ref{fig:total_facts}, it varies significantly across LLMs and note types. The red line indicates the total sentence count per note type.

\begin{figure*}[!t]
    \centering
    {\includegraphics[trim={0 5.5cm 0 5cm},clip,width=\textwidth]{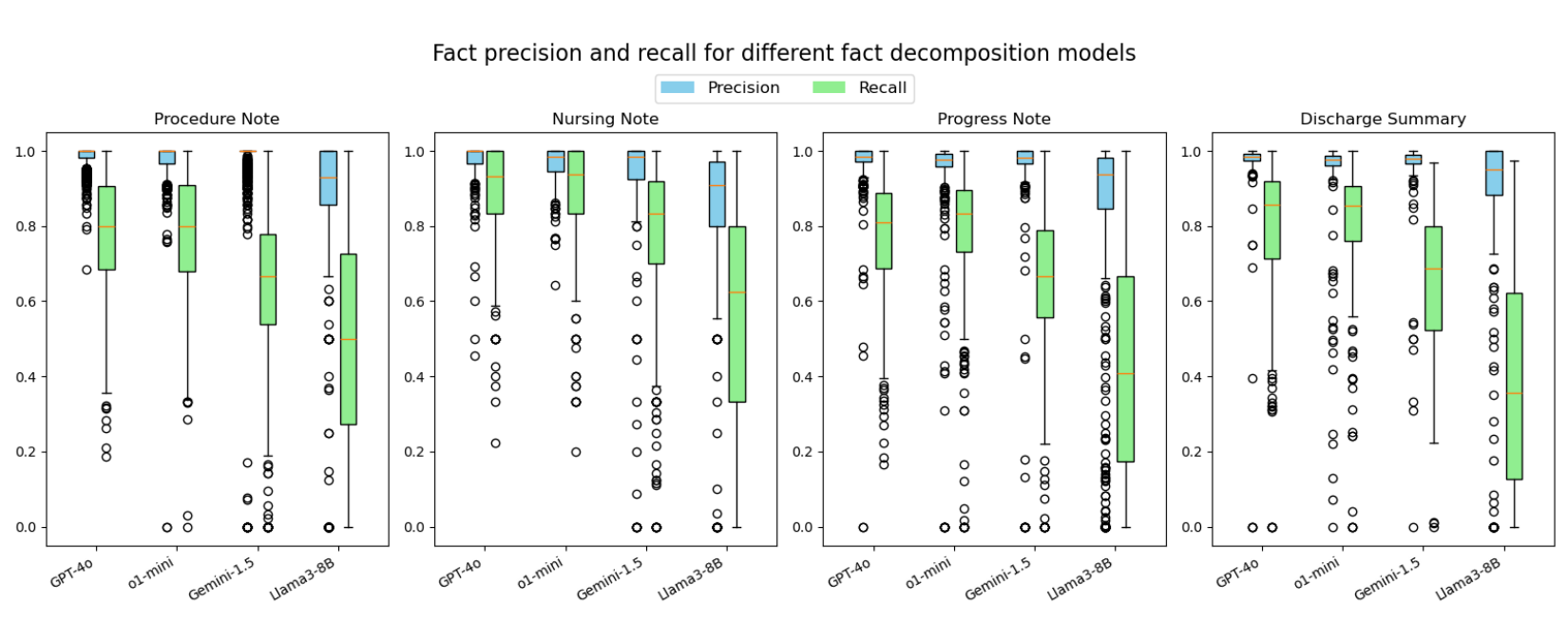}}
    \caption{Comparison of fact precision and unweighted fact recall distributions across documents for each note type and fact decomposition model.}
    \label{fig:prec_rec_variance}
    \vspace{-1em}
\end{figure*}

\subsection{Fact Precision and Fact Recall}

In Figure~\ref{fig:prec_rec_variance}, we compare the distributions of fact precision and unweighted fact recall across documents for each note type and fact decomposition model. Across all note types and models, fact recall exhibits a much wider distribution than fact precision, indicating a higher variability in the model's ability to generate comprehensive fact decompositions. This variability in recall suggests that while some documents achieve relatively high recall, many documents leave significant portions of the original information from the source notes unaccounted for missing facts in the fact decompositions.

In contrast, fact precision is generally higher and shows tighter distributions across all models and note types. This finding suggests that, for the facts that are generated, they are largely accurate and correctly entailed from the source document. 


\subsection{Entailment Model Validation}

In table~\ref{apx:entail_valid}, we present our results of GPT-4o (our evaluator model) on the human evaluation entailment dataset.

\begin{table*}[t]
\centering
\small
\begin{tabular}{ccccccc} 
\toprule
\textbf{Sensitivity} & \textbf{Specificity} & \textbf{PPV} & \textbf{NPV} & \textbf{F1} & \textbf{Accuracy} & \textbf{N} \\ 
\midrule
96.1 & 64.7 & 86.3 & 87.8 & 90.9 & 86.7 & 1036 \\ 
\bottomrule
\end{tabular}
\caption{Performance of GPT-4o on the 1,036 entailment pairs labeled by clinical experts.}
\label{apx:entail_valid}
\vspace{-1em}
\end{table*}

\section{Qualitative Evaluation}
\label{apx:other_qual_results}

\begin{figure*}[h]
    \centering
    {\includegraphics[trim={0 3.5cm 0 4cm},clip,width=\textwidth]{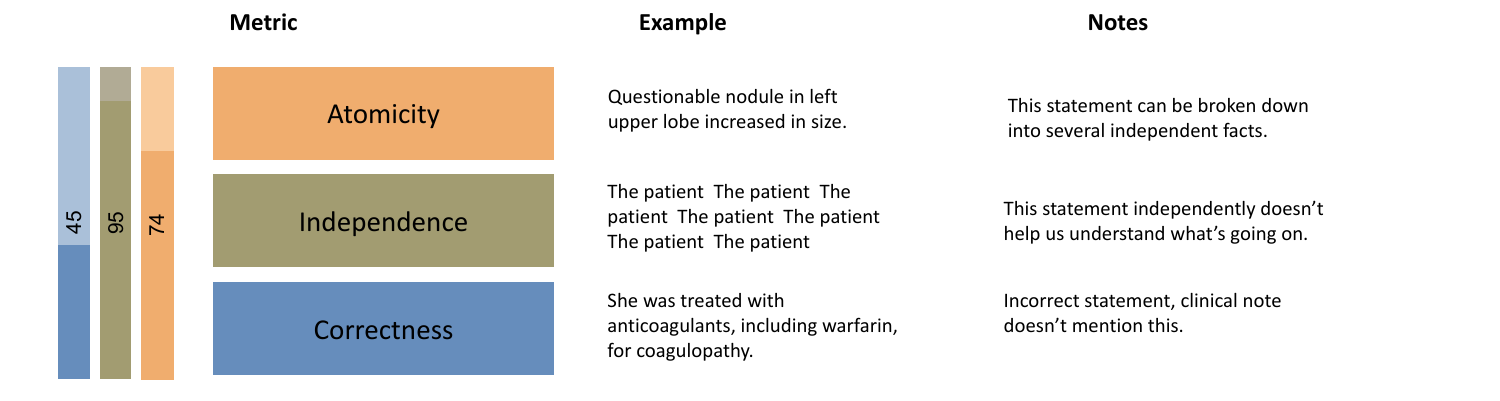}}
    \caption{Overview of in-depth qualitative review with \textbf{Llama3-8B} on the fact decomposition on twenty randomly-selected examples from FactEHR. We report \textbf{percentage} of correct, independent and atomic facts as annotated by medical expert.} 
    \label{fig:qual_eval_llama3}

\end{figure*}

\begin{figure*}[h]
    \centering
    {\includegraphics[trim={0 3.5cm 0 4cm},clip,width=\textwidth]{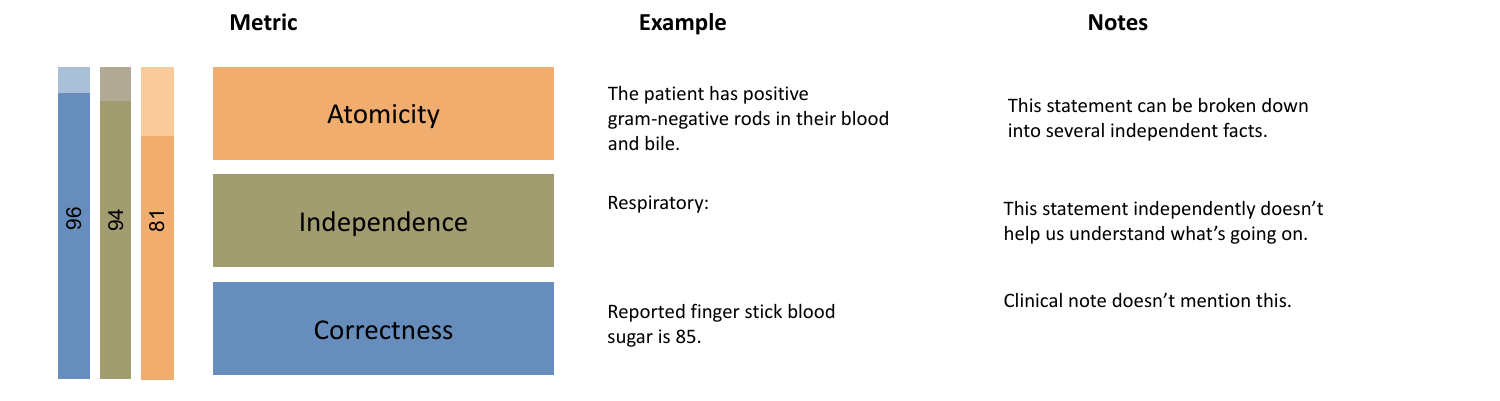}}
    \caption{Overview of in-depth qualitative review with \textbf{Gemini} on the fact decomposition on twenty randomly-selected examples from FactEHR. We report \textbf{percentage} of correct, independent and atomic facts as annotated by medical expert.} 
    \label{fig:qual_eval_gemini}
\end{figure*}

Expert reviewers assessed overall completeness of fact decompositions and the correctness, independence, and atomicity of individual facts. Detailed results for Llama3-8B and Gemini are shown in Figure~\ref{fig:qual_eval_llama3}
and Figure~\ref{fig:qual_eval_gemini} respectively \footnote{We did not have a comprehensive qualitative evaluation done for o1-mini due to time constraints.}.
 
\section{Cost}
We report API costs for generating fact decomposition of our 2,168 notes and generating entailment labels for 987,266 entailment pairs in Appendix Table \ref{tab:cost_calculations}.

\begin{table*}[ht]
\centering
\begin{tabular}{llc} 
\toprule
\textbf{Task}            & \textbf{Model}           & \textbf{ API cost (USD)}              \\ 
\midrule
Entailment               & GPT-4o                  & 10,162.72                        \\ 
Fact decomposition       & GPT-4o                  & 264.26                           \\ 
Fact decomposition       & O1-mini                 & 317.11                           \\ 
Fact decomposition       & Gemini 1.5 flash        & 5.20                             \\ 
\bottomrule
\end{tabular}
\caption{API costs for generating fact decomposition of our 2,168 notes and generating entailment labels for 987,266 entailment pairs.}
\label{tab:cost_calculations}
\end{table*}

\section{Prompts}
\label{apx:prompts}

\begin{figure*}[h]
\centering
  \includegraphics[scale=0.39]
  {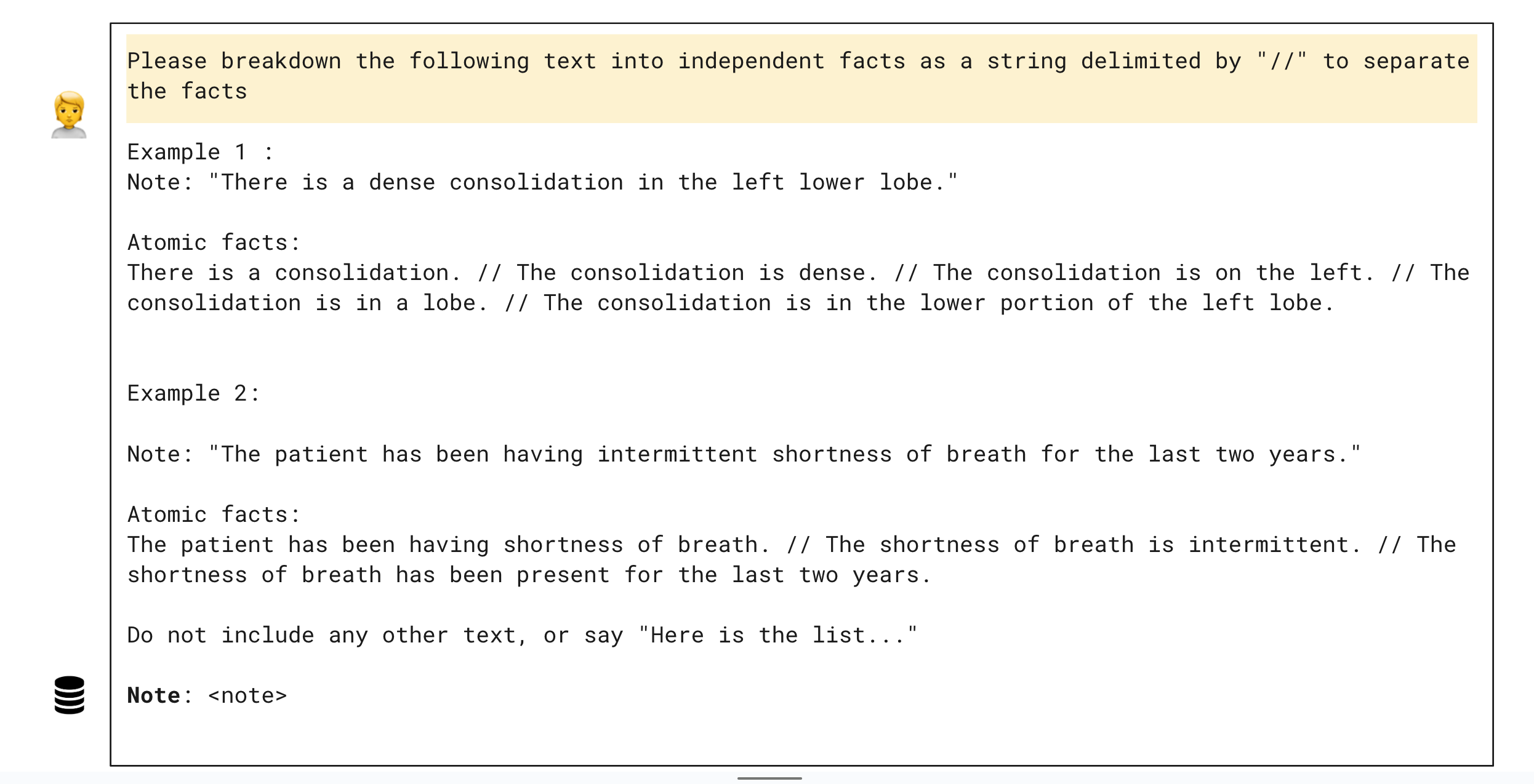}
  \caption{Prompt for fact decomposition.}
  \label{apx:fact_decomp_prompt}
\end{figure*}

We use two different prompts decribed as follows:

\paragraph{Fact Decomposition:} We adopt the prompt from \citep{min2023factscore}, and include two in-context examples of fact decompositions of clinical notes written by a medical expert, illustrated in figure~\ref{apx:fact_decomp_prompt}. The LLM is instructed to output independent facts as a delimited string. This approach resulted in fewer parsing errors than other methods. 

\paragraph{Entailment Evaluation}

\begin{figure*}[h]
\centering
  \includegraphics[scale=0.4]
  {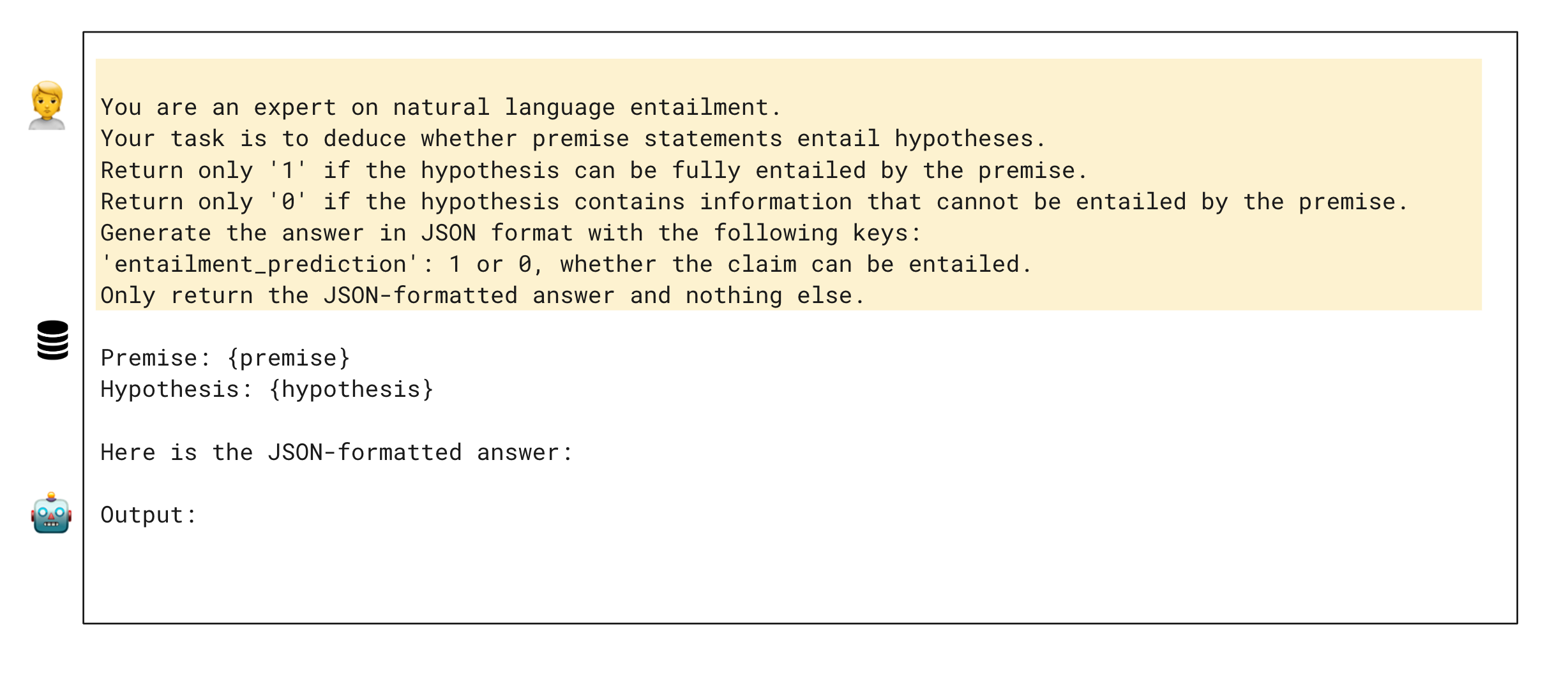}
  \caption{Prompt for entailment evaluation with GPT-4o.}
  \label{apx:entailment_prompt}
\end{figure*}

We adopt the prompt from \citep{xie-etal-2024-doclens}, instructs the LLM to output a binary 0/1 indicator of entailment in JSON format, as illustrated in figure~\ref{apx:entailment_prompt}. This prompt was preferred over those requesting rationales or chain-of-thought reasoning due to its efficiency in minimizing output tokens, reducing inference times, and lowering computational costs, especially given the large number of premise-hypothesis pairs in the FactEHR dataset.

\end{document}